\documentclass{article}

\usepackage[final]{neurips_2025}

\usepackage[utf8]{inputenc} 
\usepackage[T1]{fontenc}    
\usepackage{hyperref}       
\usepackage{url}            
\usepackage{booktabs}       
\usepackage{amsfonts}       
\usepackage{nicefrac}       
\usepackage{microtype}      
\usepackage{xcolor}         

\usepackage{amsmath}
\usepackage{mathtools}
\usepackage{appendix}
\usepackage{paralist}
\usepackage{algorithm}
\usepackage{algorithmic}

\usepackage{graphicx}
\usepackage{subfigure}

\usepackage{amsmath}
\usepackage{amssymb}
\usepackage{mathtools}
\usepackage{amsthm}
\usepackage{makecell}
\usepackage{multirow}
\usepackage{algorithm}
\usepackage{algorithmic}
\usepackage{enumitem}
\usepackage{subcaption}
\usepackage{appendix}
\usepackage{paralist}
\usepackage{wrapfig}
\usepackage{tabularx}
\usepackage{longtable}
\usepackage{ltablex}

\usepackage{fancyvrb}
\usepackage{xcolor}

\definecolor{Gray}{gray}{0.85}
\definecolor{Redo}{rgb}{0.95,0.69,0.51}
\definecolor{LightCyan}{rgb}{0.88,1,1}
\usepackage{afterpage}
\usepackage{pifont}
\usepackage{listings}

\lstdefinestyle{lfonts}{
  basicstyle   = \footnotesize\ttfamily,
  stringstyle  = \color{purple},
  keywordstyle = \color{blue!90!black}\bfseries,
  commentstyle = \color{green!60!black},
}
\lstdefinestyle{lnumbers}{
  numberstyle = \tiny,
  numbersep   = 1em,
  firstnumber = 1,
  stepnumber  = 1,
}
\lstdefinestyle{llayout}{
  breaklines       = true,
  tabsize          = 2,
  columns          = flexible,
}
\lstdefinestyle{lgeometry}{
  xleftmargin      = 20pt,
  xrightmargin     = 0pt,
  frame            = tb,
  framesep         = \fboxsep,
  framexleftmargin = 10pt,
}
\lstdefinestyle{lgeneral}{
  style = lfonts,
  style = lnumbers,
  style = llayout,
  style = lgeometry,
}
\lstdefinestyle{python}{
    language = {Python},
    style    = lgeneral,
}

\newtheorem{theorem}{Theorem}
\newtheorem{proposition}{Proposition}

\theoremstyle{definition}

\def \y         {\mathbf{y}}
\def \x         {\mathbf{x}}

\def \z         {\mathbf{z}}

\def \H         {\mathcal{H}}

\def \E         {\mathbb{E}}

\def \L         {\mathcal{L}}

\def \hte       {\hat{\theta}}

\DeclareMathOperator*{\argmin}{argmin}
\DeclareMathOperator*{\argmax}{argmax}

\def \SPIN   {\mathtt{SPIN}}
\def \TSPIN  {\mathtt{T}\mbox{-}\mathtt{SPIN}}

\def \KL     {\mathbb{D}_{\mathrm{KL}}}
\def \sft    {\textrm{SFT}}

\def \HA     {\mathtt{w}\mbox{/}\mathtt{o~H}\mbox{-}\mathtt{A}}
\def \TR     {\mathtt{T}\mbox{-}\mathtt{SPIN}_{\mathtt{ref}}}

\usepackage{tcolorbox}
\usepackage{listings}
\definecolor{bgcolor}{rgb}{0.95,0.95,0.95} 
\definecolor{titlecolor}{rgb}{1,1,1}       
\definecolor{keycolor}{rgb}{0.0,0.0,0.0}   
\definecolor{valuecolor}{rgb}{0.0,0.0,0.8} 
\definecolor{gray}{rgb}{0.4,0.4,0.4}       
\definecolor{redcolor}{rgb}{0.8,0.0,0.0}   
\definecolor{greencolor}{rgb}{0.45, 0.69, 0.39}
\definecolor{ao}{rgb}{0.0, 0.5, 0.0}

\title{Triplets Better Than Pairs: Towards Stable and Effective Self-Play Fine-Tuning for LLMs}

\author{%
    Yibo Wang\textsuperscript{\rm 1,2,3,}\thanks{Work done during the internship at Alibaba International Digital Commerce.}\ \ , Hai-Long Sun\textsuperscript{\rm 1,2,}$^*$, \
    Qing-Guo Chen\textsuperscript{\rm 3}, \\
    \textbf{Zhao Xu}\textsuperscript{\rm 3}\textbf{,} \
    \textbf{Weihua Luo}\textsuperscript{\rm 3}\textbf{,} \
    \textbf{Kaifu Zhang}\textsuperscript{\rm 3}\textbf{,} \
    \textbf{Lijun Zhang}\textsuperscript{\rm 1,4,2,}\thanks{Lijun Zhang is the corresponding author.} \\
    \textsuperscript{\rm 1}National Key Laboratory for Novel Software Technology, Nanjing University, Nanjing, China\\
    \textsuperscript{\rm 2}School of Artificial Intelligence, Nanjing University, Nanjing, China\\
    \textsuperscript{\rm 3}Alibaba International Digital Commerce  
    \textsuperscript{\rm 4}Pazhou Laboratory (Huangpu), Guangzhou, China  \\
    \texttt{\{wangyb, sunhl, zhanglj\}@lamda.nju.edu.cn} \\
    \texttt{\{qingguo.cqg, changgong.xz, weihua.luowh, kaifu.zkf\}@alibaba-inc.com} \\
}

\begin{document}

\maketitle

\begin{abstract}
  Recently, self-play fine-tuning ($\SPIN$) has been proposed to adapt large language models to downstream applications with scarce expert-annotated data, by iteratively generating synthetic responses from the model itself. However, $\SPIN$ is designed to optimize the current reward advantages of annotated responses over synthetic responses at hand, which may gradually vanish during iterations, leading to \textit{unstable optimization}. Moreover, the utilization of reference policy induces a \textit{misalignment} issue between the reward formulation for training and the metric for generation. To address these limitations, we propose a novel \textbf{T}riplet-based \textbf{S}elf-\textbf{P}lay f\textbf{I}ne-tu\textbf{N}ing ($\TSPIN$) method that integrates two key designs. First, beyond current advantages, $\TSPIN$ additionally incorporates historical advantages between iteratively generated responses and proto-synthetic responses produced by the initial policy. Even if the current advantages diminish, historical advantages remain effective, stabilizing the overall optimization. Second, $\TSPIN$ introduces the entropy constraint into the self-play framework, which is theoretically justified to support reference-free fine-tuning, eliminating the training-generation discrepancy. Empirical results on various tasks demonstrate not only the superior performance of $\TSPIN$ over $\SPIN$, but also its stable evolution during iterations. Remarkably, compared to supervised fine-tuning, $\TSPIN$ achieves comparable or even better performance with only $25\%$ samples, highlighting its effectiveness when faced with scarce annotated data.
\end{abstract}

\section{Introduction}
Large language models (LLMs) have demonstrated powerful generalization capabilities and profound impact across various fields \citep{NeurIPS:2022:Ouyang,ACL:2023:Huang,ArXiv:2023:Qin,ArXiv:2023:Bai,ArXiv:2023:Achiam,NeurIPS:2023:Liu,ArXiv:2024:Jiang,Others:2024:Wu,TKDE:2024:Zhao,NeurIPS:2024:Chen:TP,Others:2024:LLaVANeXT,ArXiv:2024:Hui,ArXiv:2024:Guo:DeepSeek,ArXiv:2024:Jiang:Visual,Others:2024:Anthropic,Others:2024:Google,Others:2024:OpenAI,ACL:2025:Sun,ICML:2025:Sun,ArXiv:2025:Lu,Arxiv:2025:Jiang,Arxiv:2026:Qian}. The remarkable success of LLMs is inherently linked to the utilization of vast expert annotations. Nevertheless, considerable curation costs for high-quality data pose major obstacles to downstream-task adaptation \citep{ArXiv:2024:Wang:Data_synthesis,ArXiv:2024:Tan}. Consequently, numerous efforts have been made to handling this challenge, e.g., data augmentation \citep{ArXiv:2023:Dai,ArXiv:2023:Zhu,ArXiv:2024:Guo} and model distillation \citep{ArXiv:2023:Gunasekar,ICLR:2024:Luo,ArXiv:2024:Xin}.

Recently, a promising research direction known as self-play fine-tuning has been introduced \citep{ICML:2024:Chen}, and has since been applied to various LLM fields \citep{ArXiv:2024:Yuan,ArXiv:2024:Wu,ArXiv:2024:Gao}. The core idea of self-play fine-tuning is to progressively refine the LLM by competing against itself. Formally, it can be framed as a two-player game between a main player and an opponent player, each of which corresponds to an instance policy of the same LLM. In this game, the main player aims to distinguish expert-annotated responses from generated ones, while the opponent player strives to produce high-quality synthetic responses that can deceive the main player.

Based on this framework, \citet{ICML:2024:Chen} introduce the seminal method named $\SPIN$, which is designed to optimize the relative reward advantages of annotated responses over generated ones, and employs the policy from previous iteration as a reference for reward computations. However, with the evolution of LLM, the relative advantage between two types of data gradually diminishes over time. When the relative advantage is eliminated, the optimization objective of $\SPIN$ degenerates into a constant independent of the policy, allowing \textit{any policy} to become an optimal solution and thus leading to \textit{unstable optimization}. Moreover, the incorporation of reference policy in $\SPIN$ causes a \textit{misalignment} between the reward formulation used for training and the log likelihood that guides response generation. Empirical evidences in Figure~\ref{fig:misalignment} reveal that although annotated responses yield higher rewards than generated responses, their log likelihoods become lower than synthetic samples during iterations.  
In other words, higher rewards do not necessarily lead to higher generation priority.

\begin{figure}[t]
    \centering
    \includegraphics[width=1.0\linewidth]{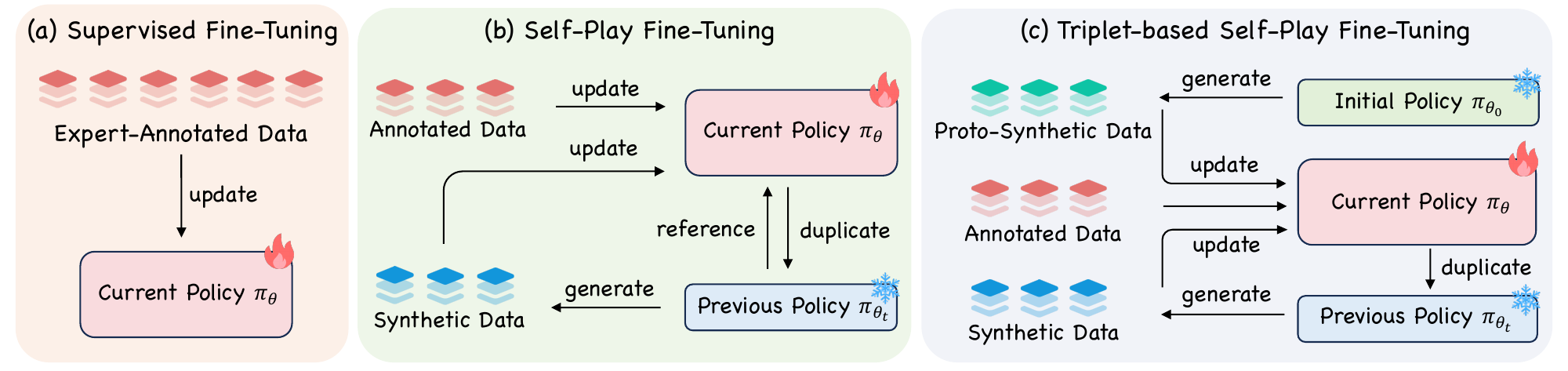}
    \caption{Comparisons of three strategies:
    (a) supervised fine-tuning requires large amounts of annotated data to train $\pi_\theta$;
    (b) self-play fine-tuning operates with limited annotated data and iteratively generated  samples, and  employs the previous policy $\pi_{\theta_t}$ as a reference for updates;
    (c) triplet-based self-play fine-tuning employs triplet inputs, i.e.,~annotated data, synthetic samples, and proto-synthetic ones from the initial policy $\pi_{\theta_0}$, and updates $\pi_\theta$ without auxiliaries from any reference policies.}
    \label{fig:comparison}
    \vspace{-1.6em}
\end{figure}

To address these limitations, we propose a new self-play fine-tuning method, termed \textbf{T}riplet-based \textbf{S}elf-\textbf{P}lay f\textbf{I}ne-tu\textbf{N}ing ($\TSPIN$). The basic intuition is to incorporate historical advantages for stable evolution, and eliminate the reference policy for training-generation alignments. Specifically, we introduce two key modifications to the self-play framework. First, the main player is required not only to distinguish between real and synthetic data, but also discriminate \textit{proto-synthetic} samples produced by the initial policy. By incorporating triplet inputs of real, synthetic and proto-synthetic samples, $\TSPIN$ optimizes current and historical advantages simultaneously. Even if the current advantage diminishes (i.e.,~synthetic samples become indistinguishable from annotated ones), the historical advantages between synthetic responses and proto-synthetic responses still exist, ensuring a stable optimization over the policy model. Second, the opponent player maximizes the confidence of synthetic responses with an \textit{entropy constraint}, where the confidence measures how likely the main player believes a response comes from the annotated data. Through theoretical analysis, we demonstrate that the incorporation of the entropy constraint removes the need for reference policy, aligning the principles of training and generation. Empirical results show that $\TSPIN$ consistently maintains higher log-likelihoods for annotated data compared to synthetic samples.

We conduct experiments on Zephyr-7B \citep{ArXiv:2023:Tunstall} and Mistral-7B \citep{ArXiv:2023:Jiang}, and select $50k$ annotated samples from Ultrachat200k \citep{ArXiv:2023:Ding} as the training set. The performance is evaluated across diverse tasks to assess multi-dimensional capabilities of LLMs, including math and logic, multi-domain knowledge, commonsense reasoning, and instruction-following. Experimental results demonstrate that $\TSPIN$ significantly improves performance, particularly in math and instruction-following, and achieves superior and stable performances during iterations compared to $\SPIN$. Furthermore, we also show that $\TSPIN$, using only $25\%$ annotated samples, can achieve comparable performance (or even better on certain tasks) to supervised fine-tuning with the full training set, highlighting the effectiveness of $\TSPIN$ in addressing the scarcity of annotated data.

\section{Preliminary}
In this section, we provide a brief overview of three fine-tuning strategies (i.e.,~supervised fine-tuning, self-play fine-tuning, and our $\TSPIN$) for LLMs. The comparisons are also illustrated in Figure~\ref{fig:comparison}.

\subsection{Supervised fine-tuning}
A common strategy to adapt a pretrained model to downstream tasks is fine-tuning it on high-quality annotated data. Specifically, given a prompt $\x$ and its corresponding annotated response $\y$, supervised fine-tuning (SFT) aims to align the model with $ (\x, \y)$. Formally, the goal of SFT is to learn a policy $\pi_{\sft}$ by maximizing the log-likelihood of high-quality annotated data:
\begin{equation}
  \label{eq:sft}
   \pi_{\sft} =   \argmax_{\pi_\theta}  \E_{\x \sim q(\cdot), \y \sim \pi_{data}(\cdot|\x)} \left[ \log \pi_\theta (\y|\x) \right],
\end{equation}
where $q(\cdot)$ and $\pi_{data}(\cdot|\x)$ denote the distributions of prompts and annotated responses, respectively. Mathematically, \eqref{eq:sft} is equivalent to the solution of minimizing the KL divergence between the target distribution $\pi_{data}(\cdot|\x)$ and the policy $\pi_\theta(\cdot|\x)$:
\begin{equation}
  \label{eq:sft:kl}
  \pi_{\sft} = \argmin_{\pi_\theta} \E_{\x \sim q(\cdot)} \left[ \KL(\pi_{data}(\cdot|\x) || \pi_\theta(\cdot|\x)) \right].
\end{equation}
It can be verified that the solution \eqref{eq:sft:kl} exactly matches the target distribution of annotated responses, i.e.,~$\pi_{\sft} (\cdot|\x) = \pi_{\text{data}}(\cdot|\x)$. Unfortunately, achieving the optimal policy typically demands extensive high-quality samples, resulting in high costs for data curation and annotation, and thereby limiting the practical applicability of supervised fine-tuning \citep{ArXiv:2024:Wang:Data_synthesis,ArXiv:2024:Tan}.

\subsection{Self-play fine-tuning}
Self-play fine-tuning has emerged as a promising way to address the scarcity of expert-annotated data when adapting LLMs to downstream tasks \citep{ICML:2024:Chen,ArXiv:2024:Yuan}. Different from SFT that runs over a fixed annotated dataset, self-play fine-tuning employs both annotated samples and self-generated responses as inputs, with the latter continuously refined as the model evolves.

Specifically, at each iteration $t+1$, the model first generates a synthetic response $\y'$ for a given prompt $\x$ by sampling from the policy $\pi_{\theta_t}(\cdot|\x)$. Then, the model utilizes the prompt $\x$ and pairwise samples (i.e., the annotated response $\y$ and the synthetic response $\y'$) to obtain the policy $\pi_{\theta_{t+1}}$. Let the reward for a response be formulated by the log ratio of the likelihood between the current policy $\pi_{\theta}$ and the previous policy $\pi_{\theta_t}$ \citep{NeurIPS:2023:Rafailov}. Then, the objective of $\SPIN$ \citep{ICML:2024:Chen} is to maximize the gap (referred to as the relative advantage) between the reward of annotated response $\y$ and that of synthetic response $\y'$:
\begin{equation}
  \label{eq:spin}
  \begin{split}
        \pi_{\theta_{t+1}} = \argmin_{\pi_\theta} \E  \left[ \ell\left( \lambda \log  \frac{\pi_\theta (\y |\x )}{ \pi_{\theta_t}(\y |\x )}   - \lambda \log   \frac{ \pi_{\theta}(\y'|\x) }{\pi_{\theta_t} (\y'|\x) } \right)   \right],
  \end{split}
\end{equation}
where $\lambda > 0$ denotes a hyper-parameter, $\ell(\cdot)$ is a convex and monotonically decreasing function, and the expectation operation is taken over the distributions $\x \sim q(\cdot), \y \sim \pi_{data}(\cdot|\x)$ and $\y' \sim \pi_{\theta_t}(\cdot|\x)$. In the following, we elaborate two major limitations of $\SPIN$: \textit{unstable optimization} during iterations and \textit{misalignment} between training and generation.

\textbf{Unstable optimization.}
The instability of $\SPIN$ arises from the inherent structure of the objective in \eqref{eq:spin}. To be precise, we consider the case where the synthetic response $\y'$ becomes identical to the annotated response $\y$ (i.e., $\y' = \y$). In this case, the advantage of $\y$ over $\y'$ vanishes, causing the objective \eqref{eq:spin} to degenerate into a constant that is independent of the policy $\pi_\theta$. As a result, \textit{any} policy $\pi_{ \theta}$ becomes optimal to \eqref{eq:spin}, leading to unstable optimization or even performance degradation.

\textbf{Misalignment between training and generation.}
Given the reference policy $\pi_{\theta_t}$ from the iteration $t$, the objective \eqref{eq:spin} is designed to optimize the reward gap between two types of responses, where the reward for a response $\hat{\y}$ is defined as $r(\x, \hat{\y}) = \lambda \log \pi_\theta (\hat{\y} |\x ) - \lambda \log \pi_{\theta_t}(\hat{\y} |\x )$. However, it is important to note that the formulation of $r(\x, \hat{\y})$ differs from the principle used for generation, i.e.,~$\log \pi_\theta(\hat{\y}|\x)$. In other words, due to the incorporation of reference policy $\pi_{\theta_t}$, a higher reward, i.e.,~$r(\x, \y) > r(\x, \z)$ for two responses $\y$ and $\z$, does not necessarily imply a higher generation probability, i.e.,~$\log \pi_\theta(\y|\x) > \log \pi_\theta(\z|\x)$ may not hold. This demonstration has been empirically verified by our experiments, as illustrated in Figure~\ref{fig:misalignment}.

\section{Triplet-based self-play fine-tuning}
\label{sec:tspin}
In this section, we introduce our triplet-based self-play fine-tuning ($\TSPIN$), a stable and reference-free method for fine-tuning LLMs. Building upon the self-play framework, our $\TSPIN$ involves two players: a main player and an opponent player, both of which are different policy instances from the same LLM. For clarity, we describe the update rules for two players separately.

\textbf{The update rule for main player.}
Recall that the goal of the main player is to distinguish among the annotated response $\y$, the synthetic response $\y'$ generated from the previous policy $\pi_{\theta_t}$ and the proto-synthetic response $\y_{0}$ produced by the initial policy $\pi_{\theta_0}$. Conceptually, the triple responses are expected to satisfy the relation $ \y \succeq \y' \succeq \y_0$, where the operator $\succeq$ indicates that the main player has greater preference for one response than in another. In other words, $\y_0$ serves as an anchor that helps to evaluate how far the current policy $\pi_{\theta_t}$ has progressed. This design is partially inspired by the schema-equilibration process in cognitive development, where knowledge builds upon and refines pre-existing cognitive anchors \citep{Others:1952:Piaget}. Formally, we are motivated by the integral probability metric (IPM) \citep{Others:1997:Muller} to establish the objective of the main player. In detail, let $c(\x, \hat{\y})$ denote the confidence with which the main player believes a response $\hat{\y}$ is sampled from the real-world data distribution $\pi_{\text{data}}(\cdot|\x)$. Then, the goal of main player is to find a confidence function 
\begin{equation}
  \label{eq:triplet:main_player}
  \begin{split}
      c_{t+1} = \argmin_{c \in \mathcal{C}} \E  \left[ \ell\left( c(\x, \y)  - c(\x, \y') \right)  + \beta \ell\left( c(\x, \y')  - c(\x, \y_{0}) \right)  \right],
  \end{split}
\end{equation}
where $\beta \geq 0$ denotes a trade-off hyper-parameter, and $\mathcal{C}$ denotes a function class that will be specified later, and $\ell(\cdot)$ denotes a monotonically decreasing function, e.g.,~the logistic function $\ell(x) = -\log \sigma(x)$ and the exponential function $\ell(x) = \exp(-x)$. The expectation operation here is taken over the distributions $\x \sim q(\cdot), \y \sim \pi_{data}(\cdot|\x), \y' \sim \pi_{\theta_t}(\cdot|\x)$ and $ \y_{0} \sim \pi_{\theta_0}(\cdot|\x)$. Given the triple responses $\y$, $\y'$, and $\y_0$, the objective in \eqref{eq:triplet:main_player} consists of two terms:

\begin{itemize}   [leftmargin=14pt]
  \vspace{-0.5mm}

  \item The first term $\ell\left( c(\x, \y) - c(\x, \y') \right)$ measures the gap between the real response $\y \sim \pi_{\text{data}}(\cdot|\x)$ and the synthetic one $\y' \sim \pi_{\theta_t}(\cdot|\x)$, reflecting the relative advantages of $\pi_{\text{data}}$ over $\pi_{\theta_t}$;

  \item The second term $\ell\left( c(\x, \y') - c(\x, \y_{0}) \right)$ compares the synthetic response $\y'$ to the proto-synthetic response $\y_0 \sim \pi_{\theta_0}(\cdot|\x)$, capturing historical advantages of $\pi_{\theta_t}$ over $\pi_{\theta_0}$ during iterations.
\end{itemize}

During the self-play process, the opponent player evolves over time, generating synthetic responses $\y'$ that progressively approximate the real response $\y$. As a result, the current advantage, i.e.,~the first term in \eqref{eq:triplet:main_player}, gradually diminishes. In contrast, since $\y_0$ remains fixed during iterations, the historical advantage, i.e.,~the second term in \eqref{eq:triplet:main_player}, remains meaningful even when the current advantage vanishes, ensuring that \eqref{eq:triplet:main_player} does not degenerate to a constant independent of $\pi_{\theta}$.

\textbf{The update rule for opponent player.} 
In this part, we suppose that the main player has learned a confidence function $c_{t+1}$ capable of effectively distinguishing between $\y$ and $\y'$. Then, we proceed to elaborate on the update rule for the opponent player. Note that the goal of the opponent player is to generate a response $\y'$ that misleads the main player to the best of its ability. Therefore, with the given confidence function $c_{t+1}$, the opponent player aims to find a policy $\pi_{\hte}$ that maximizes the expected confidence score over synthetic responses $\y'$ generated from itself:
\begin{equation}
  \label{eq:triplet:opponent_player}
  \begin{split}
      \pi_{\hte} = & \argmax_{\pi_{\theta}} \E_{\x \sim q(\cdot), \y' \sim \pi_{\theta}(\cdot|\x) } \left[ c_{t+1}(\x, \y') \right]  +   \alpha  \E_{\x \sim q(\cdot)} \left[  \H(\pi_{\theta}(\cdot|\x)) \right],
  \end{split}
\end{equation}
where $\alpha > 0$ denotes the regularization parameter, and $\H(\cdot)$ denotes the entropy constraint. In \eqref{eq:triplet:opponent_player}, the first part reflects the confidence that how the main player believes that $\y'$ is sampled from the real-world distribution $\pi_{data}$, and the second part serves to stabilize the optimization. It is worth noting that \eqref{eq:triplet:opponent_player} has a closed-form solution, as stated below.

\begin{proposition}
  \label{prop:triplet:opponent_player}
  Given a general confidence function $c_{t+1}$, the optimal policy of the opponent player in \eqref{eq:triplet:opponent_player} takes the form of
  \begin{equation}
    \pi^*(\y'|\x) = \frac{\exp(c_{t+1}(\x, \y') / \alpha)}{\sum_{\z \in \mathcal{Y}} \exp(c_{t+1}(\x, \z) / \alpha)}
    \nonumber
  \end{equation}
  where $\mathcal{Y}$ denotes the set that contains all possible responses for the prompt $\x$.
\end{proposition}
Note that due to the generality of $c_{t+1}$, the closed-form solution $\pi^*(\y'|\x)$ may not be attainable in the policy space of LLM. To resolve this issue, we need to choose a specific form of the function class $\mathcal{C}$ such that the closed-form solution becomes \textit{tractable}. Specifically, we hope to find a policy $\pi_\theta$ that satisfies $\pi_\theta(\y'|\x) = \pi^{*}(\y'|\x)$ with the confidence $c_{t+1}$ learned from a certain class $\mathcal{C}$. To this end, according to Proposition \ref{prop:triplet:opponent_player}, we can obtain $\pi_{\theta}(\y'|\x) \propto \exp(c_{t+1}(\x, \y') / \alpha )$, which implies $c_{t+1}(\x, \y') = \alpha \log \pi_{\theta}(\y'|\x)$. In other words, we can choose the function class
\begin{equation}
  \label{eq:triplet:function_class}
  \mathcal{C} = \left\{ \alpha \log \pi_{\theta}(\cdot|\x) | \theta \in \Theta \right\}
\end{equation}
where $\Theta$ denote the parameter space of LLM, and optimize \eqref{eq:triplet:main_player} to obtain the confidence $c_{t+1}(\x, \y') = \alpha \log \pi_{\theta}(\y'|\x)$ where the policy $\pi_\theta$ is exactly the optimal solution of \eqref{eq:triplet:opponent_player}. Therefore, we do not need to solve \eqref{eq:triplet:opponent_player} explicitly. Moreover, since \eqref{eq:triplet:function_class} depends solely on the policy $\pi_\theta$ and does not rely on \textit{any} reference policy, it inherently avoids the training-generation misalignment issue suffered by $\SPIN$.

\begin{algorithm}[tb]
    \caption{Triplet-based Self-Play Fine-Tuning ($\TSPIN$)}
    \label{alg:triplet}

    \textbf{Inputs}: A set of high-quality data $\{\x_i, \y_i\}_{i=1}^n$, a pretrained LLM $\pi_{\theta_0}$, and the number of iterations $T$ 

    \begin{algorithmic}[1]

        \STATE Produce the proto-synthetic response $\y_0$ with the policy $\pi_{\theta_0}$ for each prompt $\x$.

        \FOR{$t=0,1, \cdots, T-1$}
        
        \STATE For each prompt $\x$, generate one synthetic response $\y'$ by the policy $\pi_{\theta_{t}}$.

        \STATE Update the policy $\pi_{\theta_{t+1}}$ by minimizing \eqref{eq:triplet} with prompt $\x$ and triplet responses $\{\y, \y', \y_0\}$.
      
        \ENDFOR

    \end{algorithmic}
   
\end{algorithm}

\textbf{The end-to-end objective.}
We consolidate the two update rules into a unified end-to-end objective by substituting \eqref{eq:triplet:function_class} into \eqref{eq:triplet:main_player}, and obtain the loss function for $\TSPIN$ as follows:
\begin{equation}
  \label{eq:triplet}
  \begin{split}
      \L_{\TSPIN}(\theta) = &  \E  \left[ \ell\left( \alpha  \log   \pi_\theta (\y |\x )   - \alpha  \log  \pi_{\theta}(\y'|\x)   \right)  + \beta \ell\left( \alpha  \log   \pi_\theta (\y' |\x )   - \alpha  \log  \pi_{\theta}(\y_{0}|\x)   \right)  \right].
  \end{split}
\end{equation}
The expectation is taken over distributions $\x \sim q(\cdot), \y \sim \pi_{data}(\cdot|\x), \y' \sim \pi_{\theta_t}(\cdot|\x), \y_{0} \sim \pi_{\theta_0}(\cdot|\x)$. Overall, the training of $\TSPIN$ can be viewed as alternating updates between the main player and the opponent player. At each iteration $t+1$, the opponent player first generates synthetic responses $\y'$ sampled from the previous iteration policy $\pi_{\theta_{t}}$. Then, the main player takes the triplet responses $\y, \y'$ and $\y_{0}$ as inputs to obtain its confidence  $c_{t+1} = \alpha \log \pi_{\theta_{t+1}}$. After that, the opponent player chooses the policy $\pi_{\theta_{t+1}}$ for the next iteration. The detailed steps are presented in Algorithm \ref{alg:triplet}.

\textbf{Comparisons between $\TSPIN$ and $\SPIN$.}
Both methods are designed to handle the limited available data when adapting LLMs to downstream tasks, but they fundamentally differ in (i) optimization principles, (ii) implicit reward formulation, and (iii) computational overhead, as detailed below:

\begin{itemize}   [leftmargin=14pt]
  \vspace{-2.0mm}
  
  \item $\SPIN$ only optimizes the current advantage between the annotated response $ \y \sim \pi_{data}(\cdot|\x)$ and the synthetic response $\y' \sim \pi_{\theta_t}(\cdot|\x)$ in \eqref{eq:spin}, which may result in the instability issue during iterations.
  In contrast, $\TSPIN$ optimizes not only the current advantage of $\y$ over $\y'$, but also the historical advantage of $\y'$ over the proto-synthetic response $\y_0 \sim \pi_{\theta_0}(\cdot|\x)$, ensuring a stable evolution and thereby mitigating the unstable optimization;

  \vspace{-1.0mm}

  \item In $\SPIN$, the reward function for a response $\z$ is formulated as $r_{\theta}(\z|\x) = \lambda \log \pi_{\theta}(\z|\x) - \lambda \log \pi_{\theta_t}(\z|\x)$. This formulation misaligns with the generation metric (i.e.,~the log-likelihood), resulting in potential suboptimal performance. By contrast, the reward function in $\TSPIN$ (also referred to as the confidence function under the self-play framework) is defined as $r_{\theta}(\z|\x) = \alpha \log \pi_{\theta}(\z|\x)$, which matches the generation metric and thus resolves the misalignment issue;

  \vspace{-1.0mm}

  \item Compared to $\SPIN$, $\TSPIN$ incurs only a marginal increase in computational and memory costs. Specifically, $\TSPIN$ performs a one-time generation of $\y_0$ at the beginning, which is not repeated in subsequent iterations. In practice, the costs of this generation is negligible compared to the iterative training process. Moreover, while $\TSPIN$ expands each training instance from a pair of $(\y, \y')$ to a triplet of $(\y, \y', \y_0)$, the overall memory costs remains nearly unchanged, as it eliminates the need for a reference model, which is necessary in $\SPIN$. Consequently, the additional overhead of $\TSPIN$ is minimal compared to $\SPIN$.
  
\end{itemize}

\textbf{Gradient analysis.} 
To mechanistically understand the impact of $\TSPIN$ on the training process, we examine the gradient of \eqref{eq:triplet} with respect to the parameter $\theta$, as presented in the following theorem.
\begin{theorem}
  \label{theorem:gradient}
  Define $u(\pi_{\theta}) = \log \pi_{\theta}(\y|\x) - \log \pi_{\theta}(\y'|\x)$ and $v(\pi_{\theta}) = \log \pi_{\theta}(\y'|\x) - \log \pi_{\theta}(\y_0|\x)$. Then, given a monotonically decreasing function $\ell(x)$, the gradient of \eqref{eq:triplet} with respect to $\theta$ is 
  \begin{equation}
    \label{eq:triplet:gradient}
    \begin{split}
      \nabla_{\theta} \mathcal{L}_{\TSPIN}(\theta) =  \alpha \mathbb{E} \Big[ & \ell'(\alpha u(\pi_{\theta})) \cdot \left( \nabla_{\theta} \log \pi_{\theta}(\y|\x) - \nabla_{\theta} \log \pi_{\theta}(\y'|\x) \right) \\
      & + \beta \ell'(\alpha v(\pi_{\theta})) \cdot \left( \nabla_{\theta} \log \pi_{\theta}(\y'|\x) - \nabla_{\theta} \log \pi_{\theta}(\y_0|\x) \right) \Big]
    \end{split}
  \end{equation}
  where $\ell'(\cdot)$ denotes the derivative of $\ell(\cdot)$, and the expectation is taken over the distributions $\x \sim q(\cdot), \y \sim \pi_{data}(\cdot|\x), \y' \sim \pi_{\theta_t}(\cdot|\x), \y_{0} \sim \pi_{\theta_0}(\cdot|\x)$. With a specific choice of $\ell(x)$, we can derive more explicit forms of the gradient. Detailed derivation are provided in Appendix \ref{proof:gradient}.
\end{theorem}
Note that the function $\ell(x)$ is monotonically decreasing, leading to a non-positive derivative $\ell'(x) \leq 0, \forall x \in \mathbb{R}$. Therefore, during the training, the gradient in \eqref{eq:triplet:gradient} tends to increase the likelihood of $\y$ and decrease the likelihood of $\y_0$ with weights corresponding to the current advantage $u(\pi_{\theta})$ and the historical advantage $v(\pi_{\theta})$, respectively. For the synthetic response $\y'$, the update direction is influenced by both current and historical advantages, with the latter weighted by hyperparameter $\beta$.

\section{Experiments}
\label{sec:experiments}
In this section, we commence by conducting empirical evaluations of $\TSPIN$, $\SPIN$, and SFT on a set of benchmarks with $10$ different tasks. Then, we provide in-depth ablation analysis to investigate the impact of components (i.e.,~historical advantages, reward function and hyperparameters) in $\TSPIN$.

\begin{table}[t]
    \setlength{\tabcolsep}{1.5pt} 
    \renewcommand{\arraystretch}{1.1}
    \centering
    \caption{
        Performance (\%) comparisons among three fine-tuning methods, where $\SPIN$ and $\TSPIN$ are trained on $50k$ annotated data from Ultrachat200k, and SFT is trained on the complete dataset.At the average (Avg) column, we also report \textcolor{ao}{improvements} (or \textcolor{redcolor}{degradations}) over the previous iteration. Abbreviations: MMLU-Pro $\rightarrow$ MMLUP, WinoGrade $\rightarrow$ WG, Instruction Following $\rightarrow$ IF.
    }
    \vspace{0.5em}
    \resizebox{1.0 \textwidth}{!}{
        \begin{tabular}{c@{\hskip -1pt}c|ccc|ccc|ccc|c|c}
            \toprule
            \multicolumn{2}{c|}{\multirow{2}{*}{\small \textbf{Model}}} & \multicolumn{3}{c|}{ \small \textbf{Math \& Logic }} & \multicolumn{3}{c|}{\textbf{ Multi-Domain Knowledge  }} &  \multicolumn{3}{c|}{\textbf{Commonsense Reasoning }} &  \multicolumn{1}{c|}{\textbf{IF }} & \multirow{2}{*}{\textbf{\small Avg}} \\
            \cmidrule(lr){3-5} \cmidrule(lr){6-8}  \cmidrule(lr){9-11} \cmidrule(lr){12-12} 
             & & \textbf{ \small GSM8K}  & \textbf{ \small MATH} &  \textbf{ \small MUSR} & \textbf{ \small MMLU} & \textbf{ \small MMLUP} & \textbf{ \small GPQA}    & \textbf{ \small HellaSwag} & \textbf{ \small WG} & \textbf{ \small BBH}  & \textbf{ \small IFEval} & \\
            \midrule
            \multicolumn{2}{c|}{Zephyr-7B} & $25.85$ & $1.75$ & $38.92$ & $56.90$ & $28.88$ &  $28.91$ &  $82.79$  &   $74.19$ & $44.60$ & $2.76$ &    $38.56$   \\ 
            \cmidrule(lr){1-13}
            \multicolumn{2}{c|}{SFT} & $42.25$ & $3.10$ & $39.18$ &  $57.29$ & $29.19$ &  $28.28$  & $ 83.44$ & $73.16$ & $44.90$ & $19.31$  & $42.01$   \\
            \cmidrule(lr){1-13}
            \multirow{5}{*}{ \small \rotatebox{90}{  $\SPIN$ } }  &  Iter0 & $30.06$ & $4.46$ & $41.67$ & $56.25$ & $28.39$ & $29.53$ & $84.29$ & $73.56$ & $44.41$ & $8.12$ &  $40.07$   \\
             &  Iter1 & $32.47$ & $3.05$ & $39.02$ & $56.86$ & $27.88$ & $29.06$ & $83.76$ & $73.72$ & $44.59$ & $7.95$  & $39.84_{\textcolor{redcolor}{(-0.23)}}$   \\
             &  Iter2 & $36.62$ & $2.57$ & $41.27$ & $57.74$ & $28.67$ & $28.39$ & $83.75$ & $73.95$ & $43.95$ & $14.46$   & $41.14_{\textcolor{ao}{(+1.30)}}$   \\
             &  Iter3 & $33.32$ & $2.75$ & $39.41$ & $55.88$ & $27.98$ & $27.61$ & $82.64$ & $74.19$ & $43.19$ & $22.25$   & $40.92_{\textcolor{redcolor}{(-0.22)}}$   \\
             &  Iter4 & $35.54$ & $2.72$ & $38.21$ & $53.59$ & $25.95$ & $26.21$ & $83.48$ & $74.51$ & $43.10$ & $22.88$   & $40.62_{\textcolor{redcolor}{(-0.30)}}$   \\
             \cmidrule(lr){1-13}
             \multirow{5}{*}{ \small \rotatebox{90}{  $\TSPIN$  } }  &   Iter0 & $36.20$ & $3.29$ & $37.05$ & $56.42$ & $28.88$ & $28.05$ & $82.97$ & $73.48$ & $44.17$ & $7.02$  & $39.75$  \\
             &  Iter1 & $40.64$ & $3.68$ & $37.84$ &  $56.89$ & $29.13$ &  $28.92$  & $82.95$ & $73.72$ & $44.82$ & $26.98$  & $42.56_{\textcolor{ao}{(+2.81)}}$  \\
             &  Iter2 & $40.69$ & $3.63$ & $38.91$ &  $56.89$ & $29.26$ &  $29.40$  & $83.37$ & $73.64$ & $44.48$ & $27.67$  & $42.79_{\textcolor{ao}{(+0.23)}}$   \\
             &  Iter3 & $40.71$ & $3.82$ & $39.18$ &  $57.67$ & $29.37$ &  $30.12$  & $83.30$ & $73.74$ & $45.27$ & $29.15$  & $43.23_{\textcolor{ao}{(+0.44)}}$   \\
             &  Iter4 & $40.67$ & $3.84$ & $39.72$ &  $57.68$ & $29.32$ &  $30.44$  & $83.12$ & $73.80$ & $45.00$ & $31.08$  & $43.47_{\textcolor{ao}{(+0.24)}}$   \\
             \bottomrule

        \end{tabular}
        \vspace{-0.4em}
    }
    \label{tab:zephyr}
\end{table} 

\begin{figure}[t]
    \centering
    \includegraphics[width=1.0\linewidth]{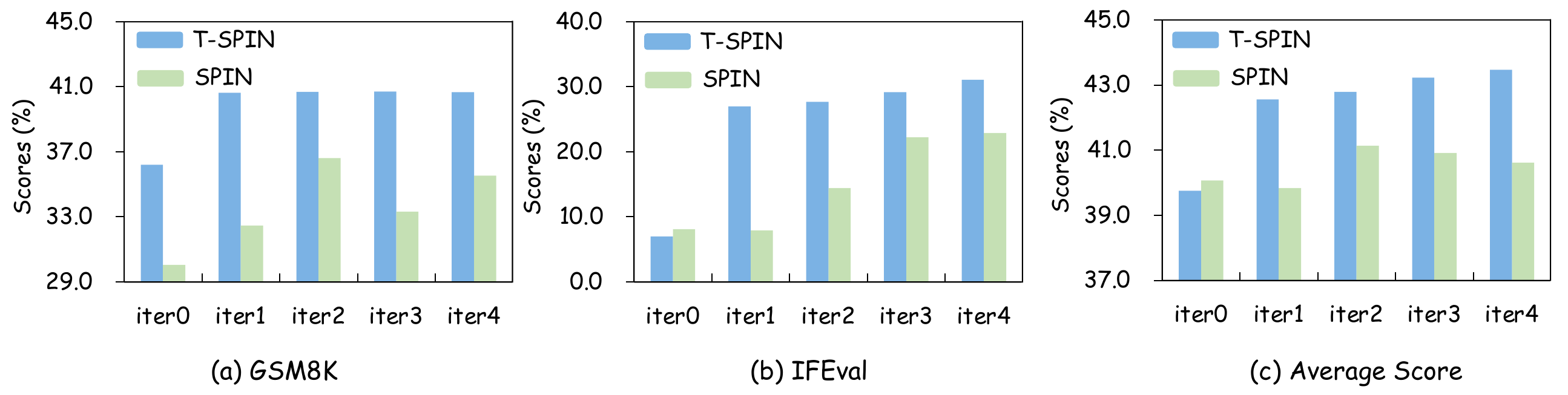}
    \caption{ 
        Performance (\%) comparisons between $\TSPIN$ and $\SPIN$ on two tasks: GSM8K and IFEval over $5$ iterations. The average scores over $10$ different tasks are also illustrated in the right panel.
    }
    \label{fig:zephyr}
    \vspace{-0.6em}
\end{figure}

\subsection{Experimental setups}
\textbf{Models and training settings.}
Following the setup in \citet{ICML:2024:Chen}, we randomly sample $50k$ prompt-response pairs from the first round of interactions in Ultrachat200k \citep{ArXiv:2023:Ding} as the training set, and choose Zephyr-7B-SFT-Full \citep{ArXiv:2023:Tunstall} and Mistral-7B-v0.1 \citep{ArXiv:2023:Jiang} as pretrained LLMs in our experiments. We employ the commonly used template: ``\verb|### Instruction: {prompt}\n\n### Response:|'' \citep{Others:2023:Taori} to construct the prompts, and set $\ell(x) = -\log\sigma(x)$ by the logistic loss function in the objectives of $\TSPIN$ and $\SPIN$. The training process adheres to the iterative framework described in Section~\ref{sec:tspin}. At each iteration, we first employ the latest policy to generate synthetic responses, which are then combined with annotated responses as the training data for the next iteration. In total, there are $50k$ synthetic responses and $50k$ annotated responses in each iteration. For $\TSPIN$, we additionally incorporate $50k$ proto-synthetic responses produced by the initial policy, into the training set.

\textbf{Evaluation benchmarks.}
We evaluate the performance across diverse tasks, each designed to assess specific capabilities of LLMs.These tasks include mathematical and logical reasoning with GSM8K \citep{ArXiv:2021:Cobbe}, MATH \citep{ArXiv:2021:Hendrycks}, MUSR \citep{ArXiv:2024:Sprague}; multi-domain knowledge understanding through MMLU \citep{ICLR:2021:Hendrycks}, MMLU-Pro \citep{ArXiv:2024:Wang:MMLU}, GPQA \citep{ArXiv:2023:Rein}; commonsense reasoning by HellaSwag \citep{ACL:2019:Zellers}, Winogrande \citep{Others:2021:Sakaguchi} and BBH \citep{ArXiv:2022:Suzgun}; and instruction-following via IFEval \citep{ArXiv:2023:Zhou:Instruction}. For each task, we follow the standard evaluation protocols in the literature. For instance, we utilize 3-shot examples and normalized accuracy as the metric in BBH. More detailed introductions about the evaluation benchmarks can be found in Appendix~\ref{sec:appendix:benchmarks}.

\begin{figure}[t]
    \centering
    \includegraphics[width=1.0\linewidth]{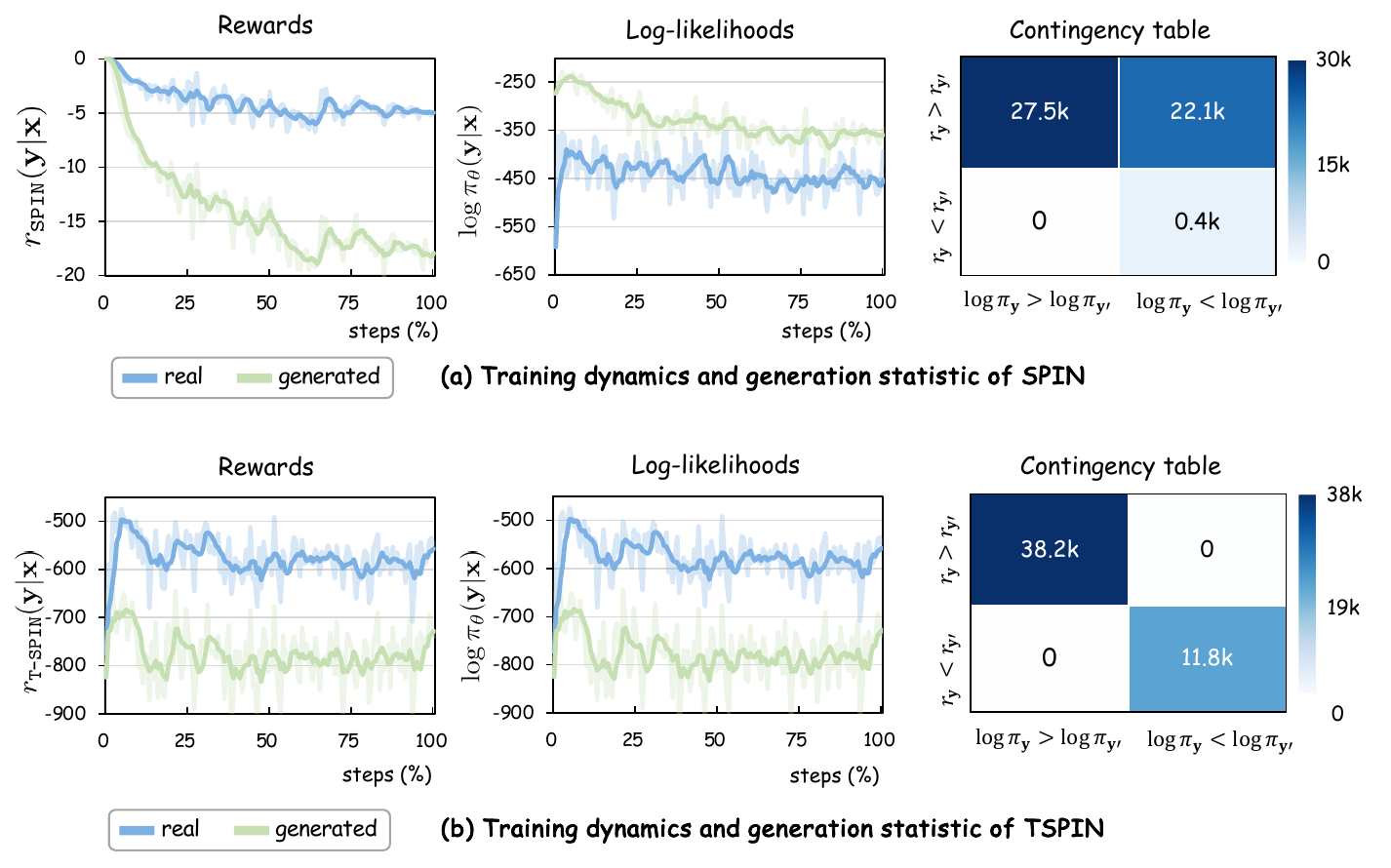}
    \caption{ 
        Comparisons between $\SPIN$ and $\TSPIN$ at iteration $1$: (a) Training dynamics (including rewards and log-likelihoods of $\y$ and $\y'$) and generation statistic (i.e.,~associations between rewards and log-likelihoods) measured on the training set for iteration $1$ of $\SPIN$; (b) Training dynamics and generation statistic of $\TSPIN$. For brevity, we denote $r_\y = r(\y|\x)$ and $\log \pi_{\y} = \log \pi (\y|\x)$.
    }
    \label{fig:misalignment}
    \vspace{-0.9em}
\end{figure}

\subsection{Experimental results}
\label{sec:main_results}
\textbf{Comparisons to $\SPIN$ and SFT.}
We first compare $\TSPIN$ with $\SPIN$ and SFT on multiple tasks, where two self-play methods are trained with only $50k$ annotated data, while SFT is trained with $200k$ annotated data. Table~\ref{tab:zephyr} summarizes the results on Zephyr-7B, and the results on Mistral-7B are deferred to the Appendix~\ref{sec:appendix:mistral}. The results demonstrate that $\TSPIN$ significantly improve the performances of pretrained models, and outperform other two fine-tuning methods. Specifically, $\TSPIN$ achieves a significant improvement in average score compared to Zephyr-7B, particularly excelling in GSM8K and IFEval tasks with $14.82$-point and $28.32$-point increases respectively, which is also illustrated in Figure~\ref{fig:zephyr}(a) and Figure~\ref{fig:zephyr}(b). Moreover, from the results shown in Figure~\ref{fig:zephyr}(c), we also observe that $\SPIN$ suffers from performance instability over iterations, reaching the peak performance at the third iteration and then suffering the degradation at the subsequent iterations. In contrast, $\TSPIN$ demonstrates stable evolutions over all iterations. Compared to SFT that utilizes the complete annotated set as training data, $\TSPIN$ with only $50k$ annotated data achieves comparable (or better) performances, highlighting the effectiveness of $\TSPIN$ when facing the data scarcity issue.

\textbf{Aligning training rewards and generation log-likelihoods.}
We then show that $\TSPIN$ is able to alleviate the misalignment issue suffered by $\SPIN$. We plot the training dynamics and generation statistics of $\SPIN$ and $\TSPIN$ on Zephyr-7B at the iteration $1$ in Figure~\ref{fig:misalignment}(a) and Figure~\ref{fig:misalignment}(b), respectively. We observe that in $\SPIN$, annotated responses yield higher rewards than synthetic responses, but their log-likelihoods are lower than those of synthetic responses, shown in the left and middle panels of Figure~\ref{fig:misalignment}(a). The reason lies in that the reward function of $\SPIN$ incorporates both the current policy $\pi_\theta$ and the reference policy $\pi_{\theta_t}$ from the previous iteration, expressed as $r_{\SPIN}(\x, \z) = \lambda \log \pi_\theta (\z |\x ) - \lambda \log \pi_{\theta_t}(\z |\x )$ for a response $\z$. This formulation differs from the generation principle that solely relies on $\log \pi_\theta (\z |\x )$. Consequently, responses with higher reward does not necessarily enjoy higher generation probability. To further illustrate this phenomenon, we also analyze the generation statistics of rewards and log-likelihoods at the right panel of Figure~\ref{fig:misalignment}(a). The results indicate that nearly half of the samples exhibit higher rewards but lower log-likelihoods. By contrast, $\TSPIN$ effectively mitigates the mismatch issue by aligning the reward formulation for training and log-likelihood used for generation, i.e.,~$r_{\TSPIN}(\x, \z) = \alpha \log \pi_\theta (\z |\x )$ where $\alpha = 1$ in this experiment. Empirical evidence, shown in the left and middle panels of Figure~\ref{fig:misalignment}(b), demonstrates that both the rewards and log-likelihoods of real responses are consistently higher than those of generated responses. We also examine the generation statistic of two types of responses in terms of rewards and log-likelihoods, with similar trends observed at the right panel of Figure~\ref{fig:misalignment}(b).

\begin{figure}[t]
    \centering
    \includegraphics[width=1.0\linewidth]{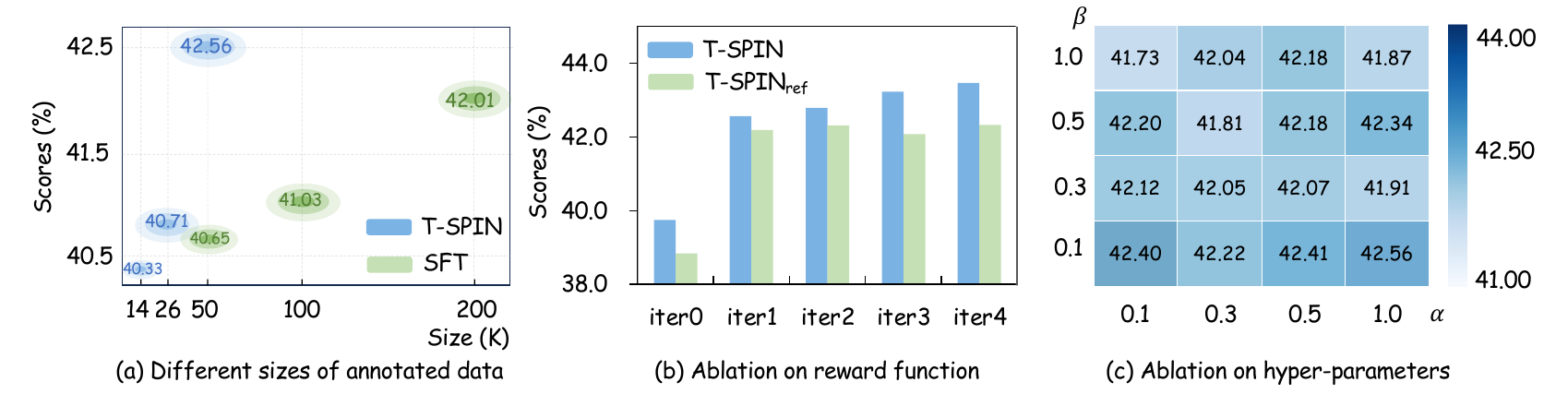}
    \caption{ 
        Performance comparisons over different settings: (a) comparison between $\TSPIN$ and SFT with varying amounts of annotated data; (b) performances of $\TSPIN$ versus the variant without historical advantage ($\HA$); (c) robustness analysis with respect to hyperparameters $\alpha$ and $\beta$.
    }
    \label{fig:ablation}
    \vspace{-0.8em}
\end{figure}

\textbf{Fine-tuning with less annotated data.}
We now demonstrate that $\TSPIN$ can achieve comparable or even superior performances to SFT, while using less annotated training data. To validate this claim, we conduct experiments with different sizes of annotated data. Specifically, we construct subsets of $14k$, $26k$, $50k$, $100k$, and $200k$ samples from Ultrachat200k \citep{ArXiv:2023:Ding}, where each larger subset encompasses all smaller ones. For comparison, we train $\TSPIN$ on Zephyr-7B using the smaller subsets ($14k$, $26k$, and $50k$ samples) and SFT using the larger subsets ($50k$, $100k$ and $200k$ samples). We report the performances of $\TSPIN$ after $2$ iterations and SFT after $2$ consecutive epochs. Figure~\ref{fig:ablation}(a) illustrates the empirical results, from which we can observe that: (i) both fine-tuning methods show improved performance with the increased annotated data size; (ii) with the same amount of annotated data (i.e.,~$50k$), $\TSPIN$ achieves a superior average score of $42.56\%$ compared to $40.65\%$ of SFT; (iii) $\TSPIN$ using only $50k$ annotated responses attains $42.56\%$ average score, slightly outperforming $42.01\%$ of SFT trained with $200k$ annotated data.

\subsection{Ablation studies}
\label{sec:ablation}
In this part, we conduct ablation studies on Zephyr-7B to investigate the impact of each component (i.e.,~historical advantages, reference-free reward function, and hyper-parameters) in $\TSPIN$.

\begin{table}[t]
    \setlength{\tabcolsep}{1.5pt} 
    \renewcommand{\arraystretch}{1.1}
    \centering
    \caption{
        Performance  (\%) comparisons between $\TSPIN$ and $\HA$ on Zephyr-7B. At the average (Avg) column, we also report  the performance \textcolor{ao}{improvements}  (or \textcolor{redcolor}{degradations})  over the previous iteration. Abbreviations:  MMLU-Pro $\rightarrow$ MMLUP, WinoGrade  $\rightarrow$ WG, Instruction Following $\rightarrow$ IF.
    }
    \vspace{0.5em}
    \resizebox{1.0 \textwidth}{!}{
        \begin{tabular}{c@{\hskip -1pt}c|ccc|ccc|ccc|c|c}
            \toprule
            \multicolumn{2}{c|}{\multirow{2}{*}{\small \textbf{Methods}}} & \multicolumn{3}{c|}{ \small \textbf{Math \& Logic     }} & \multicolumn{3}{c|}{\textbf{ Multi-Domain Knowledge  }} &  \multicolumn{3}{c|}{\textbf{Commonsense Reasoning }} &  \multicolumn{1}{c|}{\textbf{IF }} & \multirow{2}{*}{\textbf{\small Avg}} \\
            \cmidrule(lr){3-8} \cmidrule(lr){9-12}
             & &    \textbf{ \small GSM8K}  & \textbf{ \small MATH} &  \textbf{ \small MUSR} & \textbf{ \small MMLU}  & \textbf{ \small MMLUP} & \textbf{ \small GPQA}    & \textbf{ \small HellaSwag} & \textbf{ \small WG} & \textbf{ \small BBH}  & \textbf{ \small IFEval} & \\
            \midrule
            \multicolumn{2}{c|}{Iter0} & $36.20$ & $3.29$ & $37.05$ & $56.42$ & $28.88$ & $28.05$ & $82.97$ & $73.48$ & $44.17$ & $7.02$ &  $39.75$ \\
            \cmidrule(lr){1-13}
            \multirow{4}{*}{ \small \rotatebox{90}{  $\HA$ } }   &  Iter1 &  $29.89$ & $3.33$ & $37.06$ & $57.51$ & $28.98$ & $27.35$ & $82.13$ & $73.16$ & $44.63$ & $10.46$ &  $39.45_{\textcolor{redcolor}{(-0.30)}}$  \\
             &  Iter2 &  $38.95$ & $3.19$ & $37.83$ & $56.72$ & $27.65$ & $28.19$ & $83.29$ & $73.95$ & $44.64$ & $21.87$ &  $41.63_{\textcolor{ao}{(+2.18)}}$  \\
             &  Iter3 & $41.71$ & $2.63$ & $37.59$ & $56.87$ & $27.82$ & $28.96$ & $83.28$ & $73.32$ & $44.63$ & $20.10$ &  $41.64_{\textcolor{ao}{(+0.01)}}$   \\
             &  Iter4 & $38.12$ & $3.83$ & $37.05$ & $56.96$ & $28.40$ & $28.61$ & $83.71$ & $74.11$ & $43.86$ & $23.20$ &  $41.79_{\textcolor{ao}{(+0.05)}}$   \\
             \cmidrule(lr){1-13}
             \multirow{4}{*}{ \small \rotatebox{90}{  $\TSPIN$  } } &   Iter1 & $40.64$ & $3.68$ & $37.84$ &  $56.89$ & $29.13$ &  $28.92$  & $82.95$ & $73.72$ & $44.82$ & $26.98$  & $42.56_{\textcolor{ao}{(+2.81)}}$  \\
             &  Iter2 & $40.69$ & $3.63$ & $38.91$ &  $56.89$ & $29.26$ &  $29.40$  & $83.37$ & $73.64$ & $44.48$ & $27.67$  & $42.79_{\textcolor{ao}{(+0.23)}}$   \\
             &  Iter3 & $40.71$ & $3.82$ & $39.18$ &  $57.67$ & $29.37$ &  $30.12$  & $83.30$ & $73.74$ & $45.27$ & $29.15$  & $43.23_{\textcolor{ao}{(+0.44)}}$   \\
             &  Iter4 & $40.67$ & $3.84$ & $39.72$ &  $57.68$ & $29.32$ &  $30.44$  & $83.12$ & $73.80$ & $45.00$ & $31.08$  & $43.47_{\textcolor{ao}{(+0.24)}}$   \\
             \bottomrule
        \end{tabular}
    }
    \label{tab:HA}
\end{table}

\begin{figure}[ht]
    \centering
    \includegraphics[width=1.0\linewidth]{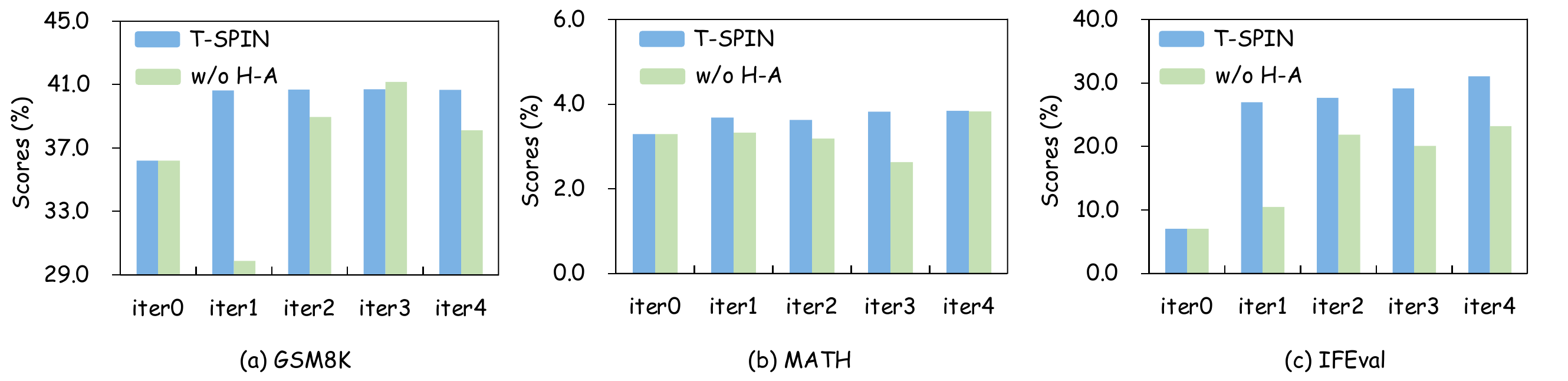}
    \caption{ 
        Performance (\%) comparisons between $\TSPIN$ and $\HA$ on GSM8K, MATH and IFEval.
    }
    \label{fig:HA}
\end{figure}

\textbf{Historical advantages.}
In $\TSPIN$, the historical advantage, i.e.,~the second term in \eqref{eq:triplet}, is introduced to address the instability issues caused by the vanishing current advantage, i.e.,~the first term in \eqref{eq:triplet}. To verify the effectiveness of historical advantages, we conduct ablation experiments by comparing $\TSPIN$ with a variant that removes the historical advantage from \eqref{eq:triplet}, referred to as $\HA$. Specifically, for $\TSPIN$, we initialize $\beta = 0$ in the first iteration, as the historical advantage is identical to the current advantage at iteration $0$, and set $\beta = 0.1$ for all subsequent iterations. For $\HA$, we fix $\beta = 0$ over all iterations. We report the performances of $\TSPIN$ and $\HA$ on all tasks in Table~\ref{tab:HA}. The average performances over different tasks reveal two key findings: (i) both $\TSPIN$ and $\HA$ are able to improve the performances, although $\HA$ suffers from a performance decline at iteration $1$; (ii) without the historical advantage, $\HA$ improves slowly and plateaus at $41.64\%$ at iteration $3$, whereas $\TSPIN$ reaches a higher performance of $42.56\%$ earlier, at iteration $1$. When we dig into the performance of each task, we find that $\HA$ suffers from unstable performance trends, as shown in Figure~\ref{fig:HA}. For example, on the task of GSM8K, $\HA$ suffers from a significant performance decline at the iteration $1$. On MATH and IFEval, the performance degradation occurs at iteration $3$. In contrast, $\TSPIN$ demonstrates consistent improvements across all tasks and maintains stability when reaching the performance peak, highlighting the effectiveness of historical advantages.

\textbf{Reference-free reward function.}
To address the training-generation misalignment issue in $\SPIN$, $\TSPIN$ introduce an entropy constraint into the self-play framework as shown in \eqref{eq:triplet:opponent_player}, from which we theoretically deliver a reference-free reward function for $\TSPIN$. To investigate its effectiveness, we compete our $\TSPIN$ with a variant, namely $\TR$, that still optimizes with triplet inputs, but uses the reward function with reference policy. The loss function of $\TR$ is given by:
\begin{equation}
    \label{eq:TR}
        \resizebox{.94\linewidth}{!}{ $
            \displaystyle
            \L_{\TR}(\theta) =  \E  \left[ \ell\left( 
            \alpha \log  \frac{\pi_\theta (\y |\x )}{ \pi_{\theta_t}(\y |\x )}   - \alpha \log   \frac{ \pi_{\theta}(\y'|\x) }{\pi_{\theta_t} (\y'|\x) }     \right)  + \beta \ell\left( \alpha \log  \frac{\pi_\theta (\y' |\x )}{ \pi_{\theta_t}(\y' |\x )}   - \alpha \log   \frac{ \pi_{\theta}(\y_0|\x) }{\pi_{\theta_t} (\y_0|\x) }      \right)  \right].
        $}
\end{equation}
We report the average performances of $\TSPIN$ and $\TR$ on all tasks in Figure~\ref{fig:ablation}(b). From the results, we observe that $\TSPIN$ demonstrates superior performance compared to $\TR$ over all iterations. This can be attributed to that using the reference-free reward function preserves the alignment between training and generation, which in turn facilitates performance improvement.

\textbf{Parameter robustness.}
$\TSPIN$ involves two hyper-parameters: (i) the trade-off parameter $\beta$ that balances the current and historical advantages, and (ii) the regularization coefficient $\alpha$ that controls the entropy penalty. To investigate the robustness of these parameters, we conduct experiments at iteration $1$ with varying $\alpha$ and $\beta$. We select values for $\alpha$ and $\beta$ from the set $ \{ 0.1, 0.3, 0.5, 1.0\}$, and report the average performances over all $10$ tasks with different parameters in Figure~\ref{fig:ablation}(c). The results demonstrate that $\TSPIN$ is relatively insensitive to a wide range of $\alpha$ and $\beta$ with diverse values. Accordingly, we use $\alpha = 1.0$ and $\beta = 0.1$ as the default values in our experiments.

\section{Related work}
\textbf{Self-play fine-tuning for LLMs.}
The self-play learning framework originates from TD-Gammon \citep{Others:1995:Tesauro}, and has produced profound impacts in various fields, such as multi-agent reinforcement learning \citep{NeurIPS:2017:Lanctot,ArXiv:2019:Muller,ICLR:2018:Bansal} and game-playing \citep{ArXiv:2017:Silver,Others:2017:Silver,NeurIPS:2017:Anthony}. The central idea of this framework is to train the model by playing against its previous version for progressive improvements. Recently, self-play fine-tuning has been introduced into the investigations for LLMs \citep{ICML:2024:Chen}, which then leads to a series of subsequent works \citep{ArXiv:2024:Ren,ArXiv:2024:Guo,ICML:2024:Yuan,ArXiv:2024:Rosset,ACL:2024:Cheng,NeurIPS:2024:Pang,ArXiv:2024:Wu,NeurIPS:2025:Wang:SPACE}. For example, \citet{ArXiv:2024:Gao} introduce the self-play mechanism into the recommendation systems to calibrate item-level predictions and suppress the bias issue, and \citet{ArXiv:2024:Yuan} incorporate the idea of self-play into diffusion models, resulting in a novel diffusion-based method. Different from the above studies that apply self-play fine-tuning to specific domains of LLMs, our work follows the seminal study of \citet{ICML:2024:Chen}, investigating the fundamental nature of self-play fine-tuning, and aims to resolve the unstable evolution and misalignment issues in the original method.

\textbf{Data generation in LLMs.}
Large language models are typically data-driven, heavily relying on the high-quality annotated data \citep{NeurIPS:2022:Ouyang,ArXiv:2024:Tan}. However, the available data is often limited for LLM training, and the costs of human expert annotation are prohibitively expensive. For this reason, recent efforts have been made to employ advanced LLMs, such as GPT series \citep{Others:2023:OpenAI,Others:2024:OpenAI,Others:2025:OpenAI:4.5}, Gemini series \citep{Others:2023:Google,Others:2024:Google,Others:2025:Google}, and Claude series \citep{Others:2024:Anthropic,Others:2024:Anthropic:3,Others:2025:Anthropic}, to generate synthetic data for LLM fine-tuning \citep{EMNLP:2023:Josifoski,ArXiv:2023:Gupta,ArXiv:2024:Long,ArXiv:2025:Qin}. It has been shown that in most domains, advanced expert LLMs are capable of generating high-quality data that can benefit others \citep{ArXiv:2023:Ubani,ArXiv:2023:Peng,ArXiv:2023:Xu:expert,ArXiv:2024:Yang,EMNLP:2024:Choi,ArXiv:2024:Zhang:TDG,ArXiv:2024:Wang:LANCE,Arxiv:2025:DeepSeek}. Compared to these studies, our self-play fine-tuning method focuses on using data generated by the current policy model to train itself to achieve the progressive performance improvements, and offers an alternative to expert-LLM-based strategies.

\section{Conclusion and discussion}
\label{sec:conclusion}
In this paper, we investigate self-play fine-tuning for LLMs and identify two issues of the existing method $\SPIN$: \textit{unstable optimization} during iterations and \textit{misalignment} between training and generation. To address these issues, we propose a novel self-play fine-tuning method named $\TSPIN$ with two key designs. First, our $\TSPIN$ simultaneously optimizes the current advantage of the annotated response $\y$ over the newly generated response $\y'$, and the historical advantage of $\y'$ over the proto-synthetic response $\y_0$. When current advantages vanish, historical advantages can still guide the fine-tuning process, avoiding the unstable optimization. Second, $\TSPIN$ incorporates the entropy constraint into the self-play framework, which eliminates the need for a reference policy in the reward computations, thus aligning the reward formulation and the generation metric. Empirically, we conduct extensive experiments and examine the performances of $\TSPIN$ on a variety of tasks. Experimental results show that $\TSPIN$ maintains stable evolution and achieves better performance than its contenders, and outperforms supervised fine-tuning by leveraging only $25\%$ annotated data.

\textbf{Limitations and future work.}
There are several promising directions for future research. First, at each iteration, $\TSPIN$ needs to re-generate synthetic responses for all prompts, although some of them may no longer contribute to further evolutions. Consequently, a carefully designed filter strategy is necessary to choose valuable samples before generation. Second, in $\TSPIN$, the expert-annotated data is fixed during iterations. But, in real-world applications, the target distribution may shift over time, and hence the fixed annotated samples may not be suitable for online fine-tuning. For this reason, self-play fine-tuning for non-stationary target distributions is another interesting direction, for which we believe that some advanced techniques from online learning may offer valuable insights \citep{Others:2016:Hazan,ArXiv:2019:Orabona,NeurIPS:2018:Zhang,NeurIPS:2024:Wang,AAAI:2024:Wang}.

\begin{ack}
    This work was partially supported by NSFC (U23A20382), and the Collaborative Innovation Center of Novel Software Technology and Industrialization.
\end{ack}

\newpage
\bibliography{ref}

\begin{thebibliography}{91}
\providecommand{\natexlab}[1]{#1}
\providecommand{\url}[1]{\texttt{#1}}
\expandafter\ifx\csname urlstyle\endcsname\relax
  \providecommand{\doi}[1]{doi: #1}\else
  \providecommand{\doi}{doi: \begingroup \urlstyle{rm}\Url}\fi

\bibitem[Achiam et~al.(2023)Achiam, Adler, Agarwal, Ahmad, Akkaya, Aleman, Almeida, Altenschmidt, Altman, Anadkat, and et~al.]{ArXiv:2023:Achiam}
J.~Achiam, S.~Adler, S.~Agarwal, L.~Ahmad, I.~Akkaya, F.~L. Aleman, D.~Almeida, J.~Altenschmidt, S.~Altman, S.~Anadkat, and et~al.
\newblock {GPT-4} technical report.
\newblock \emph{ArXiv e-prints}, arXiv:2303.08774, 2023.

\bibitem[Anthony et~al.(2017)Anthony, Tian, and Barber]{NeurIPS:2017:Anthony}
T.~Anthony, Z.~Tian, and D.~Barber.
\newblock Thinking fast and slow with deep learning and tree search.
\newblock In \emph{Advances in Neural Information Processing Systems 30}, pages 5360--5370, 2017.

\bibitem[Anthropic(2024{\natexlab{a}})]{Others:2024:Anthropic}
Anthropic.
\newblock Claude 3.5 sonnet.
\newblock \url{https://www.anthropic.com/news/claude-3-5-sonnet}, 2024{\natexlab{a}}.

\bibitem[Anthropic(2024{\natexlab{b}})]{Others:2024:Anthropic:3}
Anthropic.
\newblock Introducing the next generation of claude.
\newblock \url{https://www.anthropic.com/news/claude-3-family}, 2024{\natexlab{b}}.

\bibitem[Anthropic(2025)]{Others:2025:Anthropic}
Anthropic.
\newblock Claude 3.7 sonnet and claude code.
\newblock \url{https://www.anthropic.com/news/claude-3-7-sonnet}, 2025.

\bibitem[Bai et~al.(2023)Bai, Bai, Yang, Wang, Tan, Wang, Lin, Zhou, and Zhou]{ArXiv:2023:Bai}
J.~Bai, S.~Bai, S.~Yang, S.~Wang, S.~Tan, P.~Wang, J.~Lin, C.~Zhou, and J.~Zhou.
\newblock {Qwen-VL}: A versatile vision-language model for understanding, localization, text reading, and beyond.
\newblock \emph{ArXiv e-prints}, arXiv:2308.12966, 2023.

\bibitem[Bansal et~al.(2018)Bansal, Pachocki, Sidor, Sutskever, and Mordatch]{ICLR:2018:Bansal}
T.~Bansal, J.~Pachocki, S.~Sidor, I.~Sutskever, and I.~Mordatch.
\newblock Emergent complexity via multi-agent competition.
\newblock In \emph{Proceedings of the 6th International Conference on Learning Representations}, 2018.

\bibitem[Chen et~al.(2024{\natexlab{a}})Chen, Wang, Wu, Chen, Xu, Luo, Zhang, and Zhang]{NeurIPS:2024:Chen:TP}
S.~Chen, Y.~Wang, Y.-F. Wu, Q.-G. Chen, Z.~Xu, W.~Luo, K.~Zhang, and L.~Zhang.
\newblock Advancing tool-augmented large language models: Integrating insights from errors in inference trees.
\newblock In \emph{In Advances in Neural Information Processing Systems 37}, pages 106555--106581, 2024{\natexlab{a}}.

\bibitem[Chen et~al.(2024{\natexlab{b}})Chen, Deng, Yuan, Ji, and Gu]{ICML:2024:Chen}
Z.~Chen, Y.~Deng, H.~Yuan, K.~Ji, and Q.~Gu.
\newblock Self-play fine-tuning converts weak language models to strong language models.
\newblock In \emph{Proceedings of the 41st International Conference on Machine Learning}, pages 6621--6642, 2024{\natexlab{b}}.

\bibitem[Cheng et~al.(2024)Cheng, Yang, Li, Dai, Hu, Cao, Du, and Li]{ACL:2024:Cheng}
P.~Cheng, Y.~Yang, J.~Li, Y.~Dai, T.~Hu, P.~Cao, N.~Du, and X.~Li.
\newblock Adversarial preference optimization: Enhancing your alignment via {RM}-{LLM} game.
\newblock In \emph{Findings of the 62nd Conference of the Association for Computational Linguistics}, pages 3705--3716, 2024.

\bibitem[Choi et~al.(2024)Choi, Yun, Jin, and Kim]{EMNLP:2024:Choi}
J.~Choi, J.~Yun, K.~Jin, and Y.~Kim.
\newblock {Multi-News+}: Cost-efficient dataset cleansing via llm-based data annotation.
\newblock In \emph{Proceedings of the 2024 Conference on Empirical Methods in Natural Language Processing}, pages 15--29, 2024.

\bibitem[Cobbe et~al.(2021)Cobbe, Kosaraju, Bavarian, Chen, Jun, Kaiser, Plappert, Tworek, Hilton, Nakano, Hesse, and Schulman]{ArXiv:2021:Cobbe}
K.~Cobbe, V.~Kosaraju, M.~Bavarian, M.~Chen, H.~Jun, L.~Kaiser, M.~Plappert, J.~Tworek, J.~Hilton, R.~Nakano, C.~Hesse, and J.~Schulman.
\newblock Training verifiers to solve math word problems.
\newblock \emph{ArXiv e-prints}, arXiv:2110.14168, 2021.

\bibitem[Dai et~al.(2023)Dai, Liu, Liao, Huang, Cao, Wu, Zhao, Xu, Liu, Liu, Li, Zhu, Cai, Sun, Li, Shen, Liu, and Li]{ArXiv:2023:Dai}
H.~Dai, Z.~Liu, W.~Liao, X.~Huang, Y.~Cao, Z.~Wu, L.~Zhao, S.~Xu, W.~Liu, N.~Liu, S.~Li, D.~Zhu, H.~Cai, L.~Sun, Q.~Li, D.~Shen, T.~Liu, and X.~Li.
\newblock {AugGPT}: Leveraging {ChatGPT} for text data augmentation.
\newblock \emph{ArXiv e-prints}, arXiv:2302.13007, 2023.

\bibitem[DeepSeek-AI et~al.(2025)DeepSeek-AI, Guo, Yang, Zhang, Song, Zhang, Xu, Zhu, Ma, Wang, Bi, Zhang, Yu, Wu, Wu, Gou, Shao, Li, and et~al.]{Arxiv:2025:DeepSeek}
DeepSeek-AI, D.~Guo, D.~Yang, H.~Zhang, J.~Song, R.~Zhang, R.~Xu, Q.~Zhu, S.~Ma, P.~Wang, X.~Bi, X.~Zhang, X.~Yu, Y.~Wu, Z.~F. Wu, Z.~Gou, Z.~Shao, Z.~Li, and Z.~G. et~al.
\newblock Deepseek-r1: Incentivizing reasoning capability in llms via reinforcement learning.
\newblock \emph{ArXiv e-prints}, arXiv:2501.12948, 2025.

\bibitem[Ding et~al.(2023)Ding, Chen, Xu, Qin, Zheng, Hu, Liu, Sun, and Zhou]{ArXiv:2023:Ding}
N.~Ding, Y.~Chen, B.~Xu, Y.~Qin, Z.~Zheng, S.~Hu, Z.~Liu, M.~Sun, and B.~Zhou.
\newblock Enhancing chat language models by scaling high-quality instructional conversations.
\newblock \emph{ArXiv e-prints}, arXiv:2305.14233, 2023.

\bibitem[Gao et~al.(2024{\natexlab{a}})Gao, Chen, Yuan, Huang, Yu, and He]{ArXiv:2024:Gao}
C.~Gao, R.~Chen, S.~Yuan, K.~Huang, Y.~Yu, and X.~He.
\newblock {SPRec}: Leveraging self-play to debias preference alignment for large language model-based recommendations.
\newblock \emph{ArXiv e-prints}, arXiv:2412.09243, 2024{\natexlab{a}}.

\bibitem[Gao et~al.(2024{\natexlab{b}})Gao, Tow, Abbasi, Biderman, Black, DiPofi, Foster, Golding, Hsu, Noac'h, Li, McDonell, Muennighoff, Ociepa, Phang, Reynolds, Schoelkopf, Skowron, Sutawika, Tang, Thite, Wang, Wang, and Zou]{Others:2024:Gao}
L.~Gao, J.~Tow, B.~Abbasi, S.~Biderman, S.~Black, A.~DiPofi, C.~Foster, L.~Golding, J.~Hsu, A.~L. Noac'h, H.~Li, K.~McDonell, N.~Muennighoff, C.~Ociepa, J.~Phang, L.~Reynolds, H.~Schoelkopf, A.~Skowron, L.~Sutawika, E.~Tang, A.~Thite, B.~Wang, K.~Wang, and A.~Zou.
\newblock A framework for few-shot language model evaluation, 07 2024{\natexlab{b}}.

\bibitem[Google(2024)]{Others:2024:Google}
Google.
\newblock Our next-generation model: {Gemini} 1.5.
\newblock \url{https://blog.google/technology/ai/google-gemini-next-generation-model-february-2024}, 2024.

\bibitem[Google(2025)]{Others:2025:Google}
Google.
\newblock {Gemini 2.5}: Our most intelligent ai model.
\newblock \url{https://blog.google/technology/google-deepmind/gemini-model-thinking-updates-march-2025/}, 2025.

\bibitem[Google(2023)]{Others:2023:Google}
G.~T. Google.
\newblock {Gemini}: A family of highly capable multimodal models.
\newblock \emph{ArXiv e-prints}, arXiv:2312.11805, 2023.

\bibitem[Gugger et~al.(2022)Gugger, Debut, Wolf, Schmid, Mueller, Mangrulkar, Sun, and Bossan]{Others:2022:Gugger}
S.~Gugger, L.~Debut, T.~Wolf, P.~Schmid, Z.~Mueller, S.~Mangrulkar, M.~Sun, and B.~Bossan.
\newblock Accelerate: Training and inference at scale made simple, efficient and adaptable., 2022.

\bibitem[Gunasekar et~al.(2023)Gunasekar, Zhang, Aneja, Mendes, Giorno, Gopi, Javaheripi, Kauffmann, de~Rosa, Saarikivi, Salim, Shah, Behl, Wang, Bubeck, Eldan, Kalai, Lee, and Li]{ArXiv:2023:Gunasekar}
S.~Gunasekar, Y.~Zhang, J.~Aneja, C.~C.~T. Mendes, A.~D. Giorno, S.~Gopi, M.~Javaheripi, P.~Kauffmann, G.~de~Rosa, O.~Saarikivi, A.~Salim, S.~Shah, H.~S. Behl, X.~Wang, S.~Bubeck, R.~Eldan, A.~T. Kalai, Y.~T. Lee, and Y.~Li.
\newblock Textbooks are all you need.
\newblock \emph{ArXiv e-prints}, arXiv:2306.11644, 2023.

\bibitem[Guo et~al.(2024{\natexlab{a}})Guo, Zhu, Yang, Xie, Dong, Zhang, Chen, Bi, Wu, Li, Luo, Xiong, and Liang]{ArXiv:2024:Guo:DeepSeek}
D.~Guo, Q.~Zhu, D.~Yang, Z.~Xie, K.~Dong, W.~Zhang, G.~Chen, X.~Bi, Y.~Wu, Y.~Li, F.~Luo, Y.~Xiong, and W.~Liang.
\newblock Deepseek-coder: When the large language model meets programming -- the rise of code intelligence.
\newblock \emph{ArXiv e-prints}, arXiv:2401.14196, 2024{\natexlab{a}}.

\bibitem[Guo et~al.(2024{\natexlab{b}})Guo, Yao, Shen, Wei, Zhang, Wang, and Liu]{ArXiv:2024:Guo}
H.~Guo, Y.~Yao, W.~Shen, J.~Wei, X.~Zhang, Z.~Wang, and Y.~Liu.
\newblock Human-instruction-free llm self-alignment with limited samples.
\newblock \emph{ArXiv e-prints}, arXiv:2401.06785, 2024{\natexlab{b}}.

\bibitem[Gupta et~al.(2023)Gupta, Scaria, Anantheswaran, Verma, Parmar, Sawant, Baral, and Mishra]{ArXiv:2023:Gupta}
H.~Gupta, K.~Scaria, U.~Anantheswaran, S.~Verma, M.~Parmar, S.~A. Sawant, C.~Baral, and S.~Mishra.
\newblock {TarGEN}: Targeted data generation with large language models.
\newblock \emph{ArXiv e-prints}, arXiv:2310.17876, 2023.

\bibitem[Hazan(2016)]{Others:2016:Hazan}
E.~Hazan.
\newblock Introduction to online convex optimization.
\newblock \emph{Foundations and Trends in Optimization}, 2\penalty0 (3--4):\penalty0 157--325, 2016.

\bibitem[Hendrycks et~al.(2021{\natexlab{a}})Hendrycks, Burns, Basart, Zou, Mazeika, Song, and Steinhardt]{ICLR:2021:Hendrycks}
D.~Hendrycks, C.~Burns, S.~Basart, A.~Zou, M.~Mazeika, D.~Song, and J.~Steinhardt.
\newblock Measuring massive multitask language understanding.
\newblock In \emph{Proceedings of the 9th International Conference on Learning Representations}, 2021{\natexlab{a}}.

\bibitem[Hendrycks et~al.(2021{\natexlab{b}})Hendrycks, Burns, Kadavath, Arora, Basart, Tang, Song, and Steinhardt]{ArXiv:2021:Hendrycks}
D.~Hendrycks, C.~Burns, S.~Kadavath, A.~Arora, S.~Basart, E.~Tang, D.~Song, and J.~Steinhardt.
\newblock Measuring mathematical problem solving with the {MATH} dataset.
\newblock \emph{ArXiv e-prints}, arXiv:2103.03874, 2021{\natexlab{b}}.

\bibitem[Huang and Chang(2023)]{ACL:2023:Huang}
J.~Huang and K.~C.-C. Chang.
\newblock Towards reasoning in large language models: A survey.
\newblock In \emph{Findings of the 61st Conference of the Association for Computational Linguistics}, pages 1049--1065, 2023.

\bibitem[Hui et~al.(2024)Hui, Yang, Cui, Yang, Liu, Zhang, Liu, Zhang, Yu, Lu, Dang, Fan, Zhang, Yang, Men, Huang, Zheng, Miao, Quan, Feng, Ren, Ren, Zhou, and Lin]{ArXiv:2024:Hui}
B.~Hui, J.~Yang, Z.~Cui, J.~Yang, D.~Liu, L.~Zhang, T.~Liu, J.~Zhang, B.~Yu, K.~Lu, K.~Dang, Y.~Fan, Y.~Zhang, A.~Yang, R.~Men, F.~Huang, B.~Zheng, Y.~Miao, S.~Quan, Y.~Feng, X.~Ren, X.~Ren, J.~Zhou, and J.~Lin.
\newblock {Qwen2.5-Coder} technical report.
\newblock \emph{ArXiv e-prints}, arXiv:2409.12186, 2024.

\bibitem[Jiang et~al.(2023)Jiang, Sablayrolles, Mensch, Bamford, Chaplot, de~las Casas, Bressand, Lengyel, Lample, Saulnier, Lavaud, Lachaux, Stock, Scao, Lavril, Wang, Lacroix, and Sayed]{ArXiv:2023:Jiang}
A.~Q. Jiang, A.~Sablayrolles, A.~Mensch, C.~Bamford, D.~S. Chaplot, D.~de~las Casas, F.~Bressand, G.~Lengyel, G.~Lample, L.~Saulnier, L.~R. Lavaud, M.-A. Lachaux, P.~Stock, T.~L. Scao, T.~Lavril, T.~Wang, T.~Lacroix, and W.~E. Sayed.
\newblock Mistral 7{B}.
\newblock \emph{ArXiv e-prints}, arXiv:2310.06825, 2023.

\bibitem[Jiang et~al.(2024{\natexlab{a}})Jiang, Wang, Shen, Kim, and Kim]{ArXiv:2024:Jiang}
J.~Jiang, F.~Wang, J.~Shen, S.~Kim, and S.~Kim.
\newblock A survey on large language models for code generation.
\newblock \emph{ArXiv e-prints}, arXiv:2406.00515, 2024{\natexlab{a}}.

\bibitem[Jiang et~al.(2025)Jiang, Xia, Sun, Lu, Chen, Luo, Zhang, Zhan, and Ye]{Arxiv:2025:Jiang}
J.-P. Jiang, Y.~Xia, H.-L. Sun, S.~Lu, Q.-G. Chen, W.~Luo, K.~Zhang, D.-C. Zhan, and H.-J. Ye.
\newblock Multimodal tabular reasoning with privileged structured information.
\newblock \emph{ArXiv e-prints}, arXiv:2506.04088, 2025.

\bibitem[Jiang et~al.(2024{\natexlab{b}})Jiang, Ge, Ge, Shi, Yuan, and Shan]{ArXiv:2024:Jiang:Visual}
X.~Jiang, Y.~Ge, Y.~Ge, D.~Shi, C.~Yuan, and Y.~Shan.
\newblock Supervised fine-tuning in turn improves visual foundation models.
\newblock \emph{ArXiv e-prints}, arXiv:2401.10222, 2024{\natexlab{b}}.

\bibitem[Josifoski et~al.(2023)Josifoski, Sakota, Peyrard, and West]{EMNLP:2023:Josifoski}
M.~Josifoski, M.~Sakota, M.~Peyrard, and R.~West.
\newblock Exploiting asymmetry for synthetic training data generation: Synthie and the case of information extraction.
\newblock In \emph{Proceedings of the 2023 Conference on Empirical Methods in Natural Language Processing}, pages 1555--1574, 2023.

\bibitem[Lanctot et~al.(2017)Lanctot, Zambaldi, Gruslys, Lazaridou, Tuyls, Perolat, Silver, and Graepel]{NeurIPS:2017:Lanctot}
M.~Lanctot, V.~Zambaldi, A.~Gruslys, A.~Lazaridou, K.~Tuyls, J.~Perolat, D.~Silver, and T.~Graepel.
\newblock A unified game-theoretic approach to multiagent reinforcement learning.
\newblock In \emph{Advances in Neural Information Processing Systems 30}, pages 4190--4203, 2017.

\bibitem[Liu et~al.(2023)Liu, Li, Wu, and Lee]{NeurIPS:2023:Liu}
H.~Liu, C.~Li, Q.~Wu, and Y.~J. Lee.
\newblock Visual instruction tuning.
\newblock In \emph{Advances in Neural Information Processing Systems 36}, pages 34892--34916, 2023.

\bibitem[Liu et~al.(2024)Liu, Li, Li, Li, Zhang, Shen, and Lee]{Others:2024:LLaVANeXT}
H.~Liu, C.~Li, Y.~Li, B.~Li, Y.~Zhang, S.~Shen, and Y.~J. Lee.
\newblock {LLaVA-NeXT}: Improved reasoning, ocr, and world knowledge.
\newblock \url{https://llava-vl.github.io/blog/2024-01-30-llava-next/}, 2024.

\bibitem[Long et~al.(2024)Long, Wang, Xiao, Zhao, Ding, Chen, and Wang]{ArXiv:2024:Long}
L.~Long, R.~Wang, R.~Xiao, J.~Zhao, X.~Ding, G.~Chen, and H.~Wang.
\newblock On {LLMs}-driven synthetic data generation, curation, and evaluation: A survey.
\newblock \emph{ArXiv e-prints}, arXiv:2406.15126, 2024.

\bibitem[Lu et~al.(2025)Lu, Li, Xia, Hu, Zhao, Ma, Wei, Li, Duan, Zhao, Han, Li, Chen, Tang, Hou, Du, Zhou, Zhang, Ding, Li, Li, Hu, Gu, Yang, Wang, Sun, Wang, Sun, Huang, He, Shi, Zhang, Zheng, Jiang, Gao, Wu, Chen, Chen, Chen, Xu, Luo, and Zhang]{ArXiv:2025:Lu}
S.~Lu, Y.~Li, Y.~Xia, Y.~Hu, S.~Zhao, Y.~Ma, Z.~Wei, Y.~Li, L.~Duan, J.~Zhao, Y.~Han, H.~Li, W.~Chen, J.~Tang, C.~Hou, Z.~Du, T.~Zhou, W.~Zhang, H.~Ding, J.~Li, W.~Li, G.~Hu, Y.~Gu, S.~Yang, J.~Wang, H.~Sun, Y.~Wang, H.~Sun, J.~Huang, Y.~He, S.~Shi, W.~Zhang, G.~Zheng, J.~Jiang, S.~Gao, Y.-F. Wu, S.~Chen, Y.~Chen, Q.-G. Chen, Z.~Xu, W.~Luo, and K.~Zhang.
\newblock {Ovis2.5} technical report.
\newblock \emph{ArXiv e-prints}, arXiv:2508.11737, 2025.

\bibitem[Luo et~al.(2024)Luo, Xu, Zhao, Sun, Geng, Hu, Tao, Ma, Lin, and Jiang]{ICLR:2024:Luo}
Z.~Luo, C.~Xu, P.~Zhao, Q.~Sun, X.~Geng, W.~Hu, C.~Tao, J.~Ma, Q.~Lin, and D.~Jiang.
\newblock {WizardCoder}: Empowering code large language models with evol-instruct.
\newblock In \emph{Proceedings of the 12th International Conference on Learning Representations}, 2024.

\bibitem[Muller et~al.(2019)Muller, Omidshafiei, Rowland, Tuyls, Perolat, Liu, Hennes, Marris, Lanctot, Hughes, Wang, Lever, Heess, Graepel, and Munos]{ArXiv:2019:Muller}
P.~Muller, S.~Omidshafiei, M.~Rowland, K.~Tuyls, J.~Perolat, S.~Liu, D.~Hennes, L.~Marris, M.~Lanctot, E.~Hughes, Z.~Wang, G.~Lever, N.~Heess, T.~Graepel, and R.~Munos.
\newblock A generalized training approach for multiagent learning.
\newblock \emph{ArXiv e-prints}, arXiv:1909.12823, 2019.

\bibitem[Müller(1997)]{Others:1997:Muller}
A.~Müller.
\newblock Integral probability metrics and their generating classes of functions.
\newblock \emph{Advances in Applied Probability}, 29\penalty0 (2):\penalty0 429--443, 1997.

\bibitem[OpenAI(2023)]{Others:2023:OpenAI}
OpenAI.
\newblock {GPT-4}.
\newblock \url{https://openai.com/index/gpt-4-research/}, 2023.

\bibitem[OpenAI(2024)]{Others:2024:OpenAI}
OpenAI.
\newblock Hello {GPT-4o}.
\newblock \url{https://openai.com/index/hello-gpt-4o/}, 2024.

\bibitem[OpenAI(2025)]{Others:2025:OpenAI:4.5}
OpenAI.
\newblock Introducing {GPT-4.5}.
\newblock \url{https://openai.com/index/introducing-gpt-4-5/}, 2025.

\bibitem[Orabona(2019)]{ArXiv:2019:Orabona}
F.~Orabona.
\newblock A modern introduction to online learning.
\newblock \emph{ArXiv e-prints}, arXiv:1912.13213v6, 2019.

\bibitem[Ouyang et~al.(2022)Ouyang, Wu, Jiang, Almeida, Wainwright, Mishkin, Zhang, Agarwal, Slama, Ray, Schulman, Hilton, Kelton, Miller, Simens, Askell, Welinder, Christiano, Leike, and Lowe]{NeurIPS:2022:Ouyang}
L.~Ouyang, J.~Wu, X.~Jiang, D.~Almeida, C.~L. Wainwright, P.~Mishkin, C.~Zhang, S.~Agarwal, K.~Slama, A.~Ray, J.~Schulman, J.~Hilton, F.~Kelton, L.~Miller, M.~Simens, A.~Askell, P.~Welinder, P.~Christiano, J.~Leike, and R.~Lowe.
\newblock Training language models to follow instructions with human feedback.
\newblock In \emph{Advances in Neural Information Processing Systems 35}, pages 27730--27744, 2022.

\bibitem[Pang et~al.(2024)Pang, Yuan, He, Cho, Sukhbaatar, and Weston]{NeurIPS:2024:Pang}
R.~Y. Pang, W.~Yuan, H.~He, K.~Cho, S.~Sukhbaatar, and J.~Weston.
\newblock Iterative reasoning preference optimization.
\newblock In \emph{Advances in Neural Information Processing Systems 37}, 2024.

\bibitem[Peng et~al.(2023)Peng, Li, He, Galley, and Gao]{ArXiv:2023:Peng}
B.~Peng, C.~Li, P.~He, M.~Galley, and J.~Gao.
\newblock Instruction tuning with {GPT-4}.
\newblock \emph{ArXiv e-prints}, arXiv:2304.03277, 2023.

\bibitem[Piaget(1952)]{Others:1952:Piaget}
J.~Piaget.
\newblock The origins of intelligence in children.
\newblock \emph{International Universities Press}, 1952.

\bibitem[Qian et~al.(2026)Qian, Su, Hu, Zhang, Deng, Zhao, and Zhang]{Arxiv:2026:Qian}
Y.-Y. Qian, J.~Su, L.~Hu, P.~Zhang, Z.~Deng, P.~Zhao, and H.~Zhang.
\newblock {d3LLM}: Ultra-fast diffusion {LLM} using pseudo-trajectory distillation.
\newblock \emph{ArXiv e-prints}, arXiv:2601.07568, 2026.

\bibitem[Qin et~al.(2023)Qin, Liang, Ye, Zhu, Yan, Lu, Lin, Cong, Tang, Qian, Zhao, Hong, Tian, Xie, Zhou, Gerstein, Li, Liu, and Sun]{ArXiv:2023:Qin}
Y.~Qin, S.~Liang, Y.~Ye, K.~Zhu, L.~Yan, Y.~Lu, Y.~Lin, X.~Cong, X.~Tang, B.~Qian, S.~Zhao, L.~Hong, R.~Tian, R.~Xie, J.~Zhou, M.~Gerstein, D.~Li, Z.~Liu, and M.~Sun.
\newblock {ToolLLM}: Facilitating large language models to master 16000+ real-world {APIs}.
\newblock \emph{ArXiv e-prints}, arXiv:2307.16789, 2023.

\bibitem[Qin et~al.(2025)Qin, Dong, Zhang, Dong, Huang, Yang, Khademi, Zhang, Awadalla, Fung, Chen, Cheng, and Wei]{ArXiv:2025:Qin}
Z.~Qin, Q.~Dong, X.~Zhang, L.~Dong, X.~Huang, Z.~Yang, M.~Khademi, D.~Zhang, H.~H. Awadalla, Y.~R. Fung, W.~Chen, M.~Cheng, and F.~Wei.
\newblock Scaling laws of synthetic data for language models.
\newblock \emph{ArXiv e-prints}, arXiv:2503.19551, 2025.

\bibitem[Rafailov et~al.(2023)Rafailov, Sharma, Mitchell, Manning, Ermon, and Finn]{NeurIPS:2023:Rafailov}
R.~Rafailov, A.~Sharma, E.~Mitchell, C.~D. Manning, S.~Ermon, and C.~Finn.
\newblock Direct preference optimization: Your language model is secretly a reward model.
\newblock In \emph{Advances in Neural Information Processing Systems 36}, pages 53728--53741, 2023.

\bibitem[Rein et~al.(2023)Rein, Hou, Stickland, Petty, Pang, Dirani, Michael, and Bowman]{ArXiv:2023:Rein}
D.~Rein, B.~L. Hou, A.~C. Stickland, J.~Petty, R.~Y. Pang, J.~Dirani, J.~Michael, and S.~R. Bowman.
\newblock {GPQA}: A graduate-level google-proof {Q\&A} benchmark.
\newblock \emph{ArXiv e-prints}, arXiv:2311.12022, 2023.

\bibitem[Ren et~al.(2024)Ren, Cao, Lin, Liu, Han, Zeng, Wan, Cai, and Sun]{ArXiv:2024:Ren}
M.~Ren, B.~Cao, H.~Lin, C.~Liu, X.~Han, K.~Zeng, G.~Wan, X.~Cai, and L.~Sun.
\newblock Learning or self-aligning? rethinking instruction fine-tuning.
\newblock \emph{ArXiv e-prints}, arXiv:2402.18243, 2024.

\bibitem[Rosset et~al.(2024)Rosset, Cheng, Mitra, Santacroce, Awadallah, and Xie]{ArXiv:2024:Rosset}
C.~Rosset, C.-A. Cheng, A.~Mitra, M.~Santacroce, A.~Awadallah, and T.~Xie.
\newblock Direct nash optimization: Teaching language models to self-improve with general preferences.
\newblock \emph{ArXiv e-prints}, arXiv:2404.03715, 2024.

\bibitem[Sakaguchi et~al.(2021)Sakaguchi, Bras, Bhagavatula, and Choi]{Others:2021:Sakaguchi}
K.~Sakaguchi, R.~L. Bras, C.~Bhagavatula, and Y.~Choi.
\newblock {WinoGrande}: An adversarial winograd schema challenge at scale.
\newblock \emph{Communications of the ACM}, 64\penalty0 (9):\penalty0 99--106, 2021.

\bibitem[Scroccaro et~al.(2023)Scroccaro, Kolarijani, and Esfahani]{Others:2023:Scroccaro}
P.~Z. Scroccaro, A.~S. Kolarijani, and P.~M. Esfahani.
\newblock Adaptive composite online optimization: Predictions in static and dynamic environments.
\newblock \emph{IEEE Transactions on Automatic Control}, 68\penalty0 (5):\penalty0 2906--2921, 2023.

\bibitem[Silver et~al.(2017{\natexlab{a}})Silver, Huang, Maddison, Guez, Sifre, van~den Driessche, Schrittwieser, Antonoglou, Panneershelvam, Lanctot, Dieleman, Grewe, Nham, Kalchbrenner, Sutskever, Lillicrap, Leach, Kavukcuoglu, Graepel, and Hassabis]{ArXiv:2017:Silver}
D.~Silver, A.~Huang, C.~J. Maddison, A.~Guez, L.~Sifre, G.~van~den Driessche, J.~Schrittwieser, I.~Antonoglou, V.~Panneershelvam, M.~Lanctot, S.~Dieleman, D.~Grewe, J.~Nham, N.~Kalchbrenner, I.~Sutskever, T.~P. Lillicrap, M.~Leach, K.~Kavukcuoglu, T.~Graepel, and D.~Hassabis.
\newblock Mastering chess and shogi by self-play with a general reinforcement learning algorithm.
\newblock \emph{ArXiv e-prints}, arXiv:1712.01815, 2017{\natexlab{a}}.

\bibitem[Silver et~al.(2017{\natexlab{b}})Silver, Schrittwieser, Simonyan, Antonoglou, Huang, Guez, Hubert, Baker, Lai, Bolton, Chen, Lillicrap, Hui, Sifre, van~den Driessche, Graepel, and Hassabis]{Others:2017:Silver}
D.~Silver, J.~Schrittwieser, K.~Simonyan, I.~Antonoglou, A.~Huang, A.~Guez, T.~Hubert, L.~Baker, M.~Lai, A.~Bolton, Y.~Chen, T.~Lillicrap, F.~Hui, L.~Sifre, G.~van~den Driessche, T.~Graepel, and D.~Hassabis.
\newblock Mastering the game of go without human knowledge.
\newblock \emph{Nature}, 550\penalty0 (7676):\penalty0 354--359, 2017{\natexlab{b}}.

\bibitem[Sprague et~al.(2024)Sprague, Ye, Bostrom, Chaudhuri, and Durrett]{ArXiv:2024:Sprague}
Z.~Sprague, X.~Ye, K.~Bostrom, S.~Chaudhuri, and G.~Durrett.
\newblock {MuSR}: Testing the limits of chain-of-thought with multistep soft reasoning.
\newblock \emph{ArXiv e-prints}, arXiv:2310.16049, 2024.

\bibitem[Sun et~al.(2025{\natexlab{a}})Sun, Sun, Peng, and Ye]{ACL:2025:Sun}
H.-L. Sun, Z.~Sun, H.~Peng, and H.-J. Ye.
\newblock Mitigating visual forgetting via take-along visual conditioning for multi-modal long cot reasoning.
\newblock In \emph{Proceedings of the 63rd Conference of the Association for Computational Linguistics}, pages 5158--5171, 2025{\natexlab{a}}.

\bibitem[Sun et~al.(2025{\natexlab{b}})Sun, Zhou, Li, Lu, Yi, Chen, Xu, Luo, Zhang, and Zhan]{ICML:2025:Sun}
H.-L. Sun, D.-W. Zhou, Y.~Li, S.~Lu, C.~Yi, Q.-G. Chen, Z.~Xu, W.~Luo, K.~Zhang, and D.-C. Zhan.
\newblock {Parrot}: Multilingual visual instruction tuning.
\newblock In \emph{Proceedings of the 42nd International Conference on Machine Learning}, 2025{\natexlab{b}}.

\bibitem[Suzgun et~al.(2022)Suzgun, Scales, Schärli, Gehrmann, Tay, Chung, Chowdhery, Le, Chi, Zhou, and Wei]{ArXiv:2022:Suzgun}
M.~Suzgun, N.~Scales, N.~Schärli, S.~Gehrmann, Y.~Tay, H.~W. Chung, A.~Chowdhery, Q.~V. Le, E.~H. Chi, D.~Zhou, and J.~Wei.
\newblock Challenging {BIG-Bench} tasks and whether chain-of-thought can solve them.
\newblock \emph{ArXiv e-prints}, arXiv:2210.09261, 2022.

\bibitem[Tan et~al.(2024)Tan, Li, Wang, Beigi, Jiang, Bhattacharjee, Karami, Li, Cheng, and Liu]{ArXiv:2024:Tan}
Z.~Tan, D.~Li, S.~Wang, A.~Beigi, B.~Jiang, A.~Bhattacharjee, M.~Karami, J.~Li, L.~Cheng, and H.~Liu.
\newblock Large language models for data annotation and synthesis: A survey.
\newblock \emph{ArXiv e-prints}, arXiv:2402.13446, 2024.

\bibitem[Taori et~al.(2023)Taori, Gulrajani, Zhang, Dubois, Li, Guestrin, Liang, and Hashimoto]{Others:2023:Taori}
R.~Taori, I.~Gulrajani, T.~Zhang, Y.~Dubois, X.~Li, C.~Guestrin, P.~Liang, and T.~B. Hashimoto.
\newblock {Stanford Alpaca}: An instruction-following {LLaMA} model, 2023.

\bibitem[Tesauro(1995)]{Others:1995:Tesauro}
G.~Tesauro.
\newblock Temporal difference learning and {TD-Gammon}.
\newblock \emph{Communications of the ACM}, 38\penalty0 (3):\penalty0 58--68, 1995.

\bibitem[Tunstall et~al.(2023)Tunstall, Beeching, Lambert, Rajani, Rasul, Belkada, Huang, von Werra, Fourrier, Habib, Sarrazin, Sanseviero, Rush, and Wolf]{ArXiv:2023:Tunstall}
L.~Tunstall, E.~Beeching, N.~Lambert, N.~Rajani, K.~Rasul, Y.~Belkada, S.~Huang, L.~von Werra, C.~Fourrier, N.~Habib, N.~Sarrazin, O.~Sanseviero, A.~M. Rush, and T.~Wolf.
\newblock Zephyr: Direct distillation of {LM} alignment.
\newblock \emph{ArXiv e-prints}, arXiv:2310.16944, 2023.

\bibitem[Tunstall et~al.(2024)Tunstall, Beeching, Lambert, Rajani, Huang, Rasul, Bartolome, Rush, and Wolf]{Others:2024:Tunstall}
L.~Tunstall, E.~Beeching, N.~Lambert, N.~Rajani, S.~Huang, K.~Rasul, A.~Bartolome, A.~M. Rush, and T.~Wolf.
\newblock {The Alignment Handbook}, 2024.
\newblock URL \url{https://github.com/huggingface/alignment-handbook}.

\bibitem[Ubani et~al.(2023)Ubani, Polat, and Nielsen]{ArXiv:2023:Ubani}
S.~Ubani, S.~O. Polat, and R.~Nielsen.
\newblock {ZeroShotDataAug}: Generating and augmenting training data with chatgpt.
\newblock \emph{ArXiv e-prints}, arXiv:2304.14334, 2023.

\bibitem[Wang et~al.(2024{\natexlab{a}})Wang, Zhu, Ren, Liu, Li, Zhang, Zhang, Wu, Zhan, Liu, and Wang]{ArXiv:2024:Wang:Data_synthesis}
K.~Wang, J.~Zhu, M.~Ren, Z.~Liu, S.~Li, Z.~Zhang, C.~Zhang, X.~Wu, Q.~Zhan, Q.~Liu, and Y.~Wang.
\newblock A survey on data synthesis and augmentation for large language models.
\newblock \emph{ArXiv e-prints}, arXiv:2410.12896, 2024{\natexlab{a}}.

\bibitem[Wang et~al.(2024{\natexlab{b}})Wang, Wang, Ma, Yang, Feng, Wang, Zhang, and Song]{ArXiv:2024:Wang:LANCE}
P.~Wang, M.~Wang, Z.~Ma, X.~Yang, S.~Feng, D.~Wang, Y.~Zhang, and K.~Song.
\newblock Language models as continuous self-evolving data engineers.
\newblock \emph{ArXiv e-prints}, arXiv:2412.15151, 2024{\natexlab{b}}.

\bibitem[Wang et~al.(2024{\natexlab{c}})Wang, Chen, Jiang, Yang, Wan, and Zhang]{NeurIPS:2024:Wang}
Y.~Wang, S.~Chen, W.~Jiang, W.~Yang, Y.~Wan, and L.~Zhang.
\newblock Online composite optimization between stochastic and adversarial environments.
\newblock In \emph{In Advances in Neural Information Processing Systems 37}, pages 94808--94850, 2024{\natexlab{c}}.

\bibitem[Wang et~al.(2024{\natexlab{d}})Wang, Ma, Zhang, Ni, Chandra, Guo, Ren, Arulraj, He, Jiang, Li, Ku, Wang, Zhuang, Fan, Yue, and Chen]{ArXiv:2024:Wang:MMLU}
Y.~Wang, X.~Ma, G.~Zhang, Y.~Ni, A.~Chandra, S.~Guo, W.~Ren, A.~Arulraj, X.~He, Z.~Jiang, T.~Li, M.~Ku, K.~Wang, A.~Zhuang, R.~Fan, X.~Yue, and W.~Chen.
\newblock {MMLU-Pro}: A more robust and challenging multi-task language understanding benchmark.
\newblock \emph{ArXiv e-prints}, arXiv:2406.01574, 2024{\natexlab{d}}.

\bibitem[Wang et~al.(2024{\natexlab{e}})Wang, Yang, Jiang, Lu, Wang, Tang, Wan, and Zhang]{AAAI:2024:Wang}
Y.~Wang, W.~Yang, W.~Jiang, S.~Lu, B.~Wang, H.~Tang, Y.~Wan, and L.~Zhang.
\newblock Non-stationary projection-free online learning with dynamic and adaptive regret guarantees.
\newblock In \emph{Proceedings of the 38th AAAI Conference on Artificial Intelligence}, pages 15671--15679, 2024{\natexlab{e}}.

\bibitem[Wang et~al.(2025)Wang, Huzhang, Chen, Xu, Luo, Zhang, and Zhang]{NeurIPS:2025:Wang:SPACE}
Y.~Wang, G.~Huzhang, Q.-G. Chen, Z.~Xu, W.~Luo, K.~Zhang, and L.~Zhang.
\newblock {SPACE}: Noise contrastive estimation stabilizes self-play fine-tuning for large language models.
\newblock In \emph{Advances in Neural Information Processing Systems 38}, 2025.

\bibitem[Wu et~al.(2024{\natexlab{a}})Wu, Zheng, Qiu, Wang, Gu, Shen, Qin, Zhu, Zhu, Liu, Xiong, and Chen]{Others:2024:Wu}
L.~Wu, Z.~Zheng, Z.~Qiu, H.~Wang, H.~Gu, T.~Shen, C.~Qin, C.~Zhu, H.~Zhu, Q.~Liu, H.~Xiong, and E.~Chen.
\newblock A survey on large language models for recommendation.
\newblock \emph{World Wide Web}, 27\penalty0 (60), 2024{\natexlab{a}}.

\bibitem[Wu et~al.(2024{\natexlab{b}})Wu, Sun, Yuan, Ji, Yang, and Gu]{ArXiv:2024:Wu}
Y.~Wu, Z.~Sun, H.~Yuan, K.~Ji, Y.~Yang, and Q.~Gu.
\newblock Self-play preference optimization for language model alignment.
\newblock \emph{ArXiv e-prints}, arXiv:2405.00675, 2024{\natexlab{b}}.

\bibitem[Xin et~al.(2024)Xin, Guo, Shao, Ren, Zhu, Liu, Ruan, Li, and Liang]{ArXiv:2024:Xin}
H.~Xin, D.~Guo, Z.~Shao, Z.~Ren, Q.~Zhu, B.~Liu, C.~Ruan, W.~Li, and X.~Liang.
\newblock {DeepSeek-Prover}: Advancing theorem proving in llms through large-scale synthetic data.
\newblock \emph{ArXiv e-prints}, arXiv:2405.14333, 2024.

\bibitem[Xu et~al.(2023)Xu, Yang, Lin, Wang, Zhou, Zhang, and Mao]{ArXiv:2023:Xu:expert}
B.~Xu, A.~Yang, J.~Lin, Q.~Wang, C.~Zhou, Y.~Zhang, and Z.~Mao.
\newblock {ExpertPrompting}: Instructing large language models to be distinguished experts.
\newblock \emph{ArXiv e-prints}, arXiv:2305.14688, 2023.

\bibitem[Yang et~al.(2024)Yang, Zhang, Hui, Gao, Yu, Li, Liu, Tu, Zhou, Lin, Lu, Xue, Lin, Liu, Ren, and Zhang]{ArXiv:2024:Yang}
A.~Yang, B.~Zhang, B.~Hui, B.~Gao, B.~Yu, C.~Li, D.~Liu, J.~Tu, J.~Zhou, J.~Lin, K.~Lu, M.~Xue, R.~Lin, T.~Liu, X.~Ren, and Z.~Zhang.
\newblock {Qwen2.5}-math technical report: Toward mathematical expert model via self-improvement.
\newblock \emph{ArXiv e-prints}, arXiv:2409.12122v1, 2024.

\bibitem[Yuan et~al.(2024{\natexlab{a}})Yuan, Chen, Ji, and Gu]{ArXiv:2024:Yuan}
H.~Yuan, Z.~Chen, K.~Ji, and Q.~Gu.
\newblock Self-play fine-tuning of diffusion models for text-to-image generation.
\newblock \emph{ArXiv e-prints}, arXiv:2402.10210, 2024{\natexlab{a}}.

\bibitem[Yuan et~al.(2024{\natexlab{b}})Yuan, Pang, Cho, Li, Sukhbaatar, Xu, and Weston]{ICML:2024:Yuan}
W.~Yuan, R.~Y. Pang, K.~Cho, X.~Li, S.~Sukhbaatar, J.~Xu, and J.~Weston.
\newblock Self-rewarding language models.
\newblock In \emph{Proceedings of the 41st International Conference on Machine Learning}, pages 57905--57923, 2024{\natexlab{b}}.

\bibitem[Zellers et~al.(2019)Zellers, Holtzman, Bisk, Farhadi, and Choi]{ACL:2019:Zellers}
R.~Zellers, A.~Holtzman, Y.~Bisk, A.~Farhadi, and Y.~Choi.
\newblock {HellaSwag}: Can a machine really finish your sentence?
\newblock In \emph{Proceedings of the 57th Conference of the Association for Computational Linguistics}, pages 4791--4800, 2019.

\bibitem[Zhang et~al.(2018)Zhang, Lu, and Zhou]{NeurIPS:2018:Zhang}
L.~Zhang, S.~Lu, and Z.-H. Zhou.
\newblock Adaptive online learning in dynamic environments.
\newblock In \emph{Advances in Neural Information Processing Systems 31}, pages 1323--1333, 2018.

\bibitem[Zhang(2024)]{ArXiv:2024:Zhang:TDG}
Y.~Zhang.
\newblock Training and evaluating language models with template-based data generation.
\newblock \emph{ArXiv e-prints}, arXiv:2411.18104, 2024.

\bibitem[Zhao et~al.(2024)Zhao, Fan, Li, Liu, Mei, Wang, Wen, Wang, Zhao, Tang, and Li]{TKDE:2024:Zhao}
Z.~Zhao, W.~Fan, J.~Li, Y.~Liu, X.~Mei, Y.~Wang, Z.~Wen, F.~Wang, X.~Zhao, J.~Tang, and Q.~Li.
\newblock Recommender systems in the era of large language models ({LLMs}).
\newblock \emph{IEEE Transactions on Knowledge and Data Engineering}, 36:\penalty0 6889--6907, 2024.

\bibitem[Zhou et~al.(2023)Zhou, Lu, Mishra, Brahma, Basu, Luan, Zhou, and Hou]{ArXiv:2023:Zhou:Instruction}
J.~Zhou, T.~Lu, S.~Mishra, S.~Brahma, S.~Basu, Y.~Luan, D.~Zhou, and L.~Hou.
\newblock Instruction-following evaluation for large language models.
\newblock \emph{ArXiv e-prints}, arXiv:2311.07911, 2023.

\bibitem[Zhu et~al.(2023)Zhu, Zhang, Haq, Hui, and Tyson]{ArXiv:2023:Zhu}
Y.~Zhu, P.~Zhang, E.-U. Haq, P.~Hui, and G.~Tyson.
\newblock Can {ChatGPT} reproduce human-generated labels? a study of social computing tasks.
\newblock \emph{ArXiv e-prints}, arXiv:2304.10145, 2023.

\end{thebibliography}
\bibliographystyle{abbrvnat}

\newpage
\appendix
\section{Mathematical derivations}
In this section, we provide mathematical derivations for the propositions and theorems.

\subsection{Proof of Proposition \ref{prop:triplet:opponent_player}}
In this part, we proceed to derive the optimal solution of the following objective:
\begin{equation}
  \label{eq:triplet:opponent_player_proof:1}
  \pi_{\hte} = \argmax_{\pi_{\theta}} \E_{\x \sim q(\cdot), \y' \sim \pi_{\theta}(\cdot|\x) } \left[ c_{t+1}(\x, \y') \right] + \alpha \E_{\x \sim q(\cdot)} \left[ \H(\pi_{\theta}(\cdot|\x)) \right].
\end{equation}
under a general confidence function $c_{t+1}$ with the entropy constraint $\H(\cdot)$. The proof is inspired by \cite{NeurIPS:2023:Rafailov}, converting \eqref{eq:triplet:opponent_player_proof:1} into a KL-divergence minimization problem. Specifically, by the definition of entropy, we have:
\begin{equation}
  \label{eq:triplet:opponent_player_proof:2}
  \H(\pi_{\theta}(\cdot|\x)) = - \E_{\y' \sim \pi_{\theta}(\cdot|\x)} \left[ \log \pi_{\theta}(\y'|\x) \right].
\end{equation}
Combining \eqref{eq:triplet:opponent_player_proof:1} and \eqref{eq:triplet:opponent_player_proof:2}, we have:
\begin{equation}
  \label{eq:triplet:opponent_player_proof:3}
  \begin{split}
    & \argmax_{\pi_{\theta}} \E_{\x \sim q(\cdot), \y' \sim \pi_{\theta}(\cdot|\x) } \left[ c_{t+1}(\x, \y') \right]  +  \alpha  \E_{\x \sim q(\cdot)} \left[  \H(\pi_{\theta}(\cdot|\x)) \right] \\
    = & \argmax_{\pi_{\theta}} \E_{\x \sim q(\cdot), \y' \sim \pi_{\theta}(\cdot|\x) } \left[ c_{t+1}(\x, \y')   -  \alpha    \log \pi_{\theta}(\y'|\x) \right]\\
    =& \argmin_{\pi_{\theta}} \E_{\x \sim q(\cdot), \y' \sim \pi_{\theta}(\cdot|\x) } \left[     \log \pi_{\theta}(\y'|\x) - \frac{1}{\alpha} c_{t+1}(\x, \y') \right]\\
    =& \argmin_{\pi_{\theta}} \E_{\x \sim q(\cdot), \y' \sim \pi_{\theta}(\cdot|\x) } \left[     \log \frac{\pi_{\theta}(\y'|\x)}{\frac{1}{Z(\x)} \exp(\frac{1}{\alpha} c_{t+1}(\x, \y') )} - \log Z(\x) \right].
  \end{split}
\end{equation}
where $Z(\x) = \sum_{\z \in \mathcal{Y}} \exp(\frac{1}{\alpha} c_{t+1}(\x, \z) )$ denotes the partition function for normalization. Note that $Z(\x)$ is a function of the prompt $\x$, and independent of the policy $\pi_{\theta}$. We define
\begin{equation}
  \pi_{\theta^*}(\y|\x) = \frac{1}{Z(\x)} \exp\left(\frac{1}{\alpha} c_{t+1}(\x, \y) \right),
  \nonumber
\end{equation}
which is a valid policy, since $\pi_{\theta^*}(\y|\x) \geq 0$ for any $\y$ and $\sum_{\y} \pi_{\theta^*}(\y|\x) = 1$. Next, we can rewrite \eqref{eq:triplet:opponent_player_proof:3} as:
\begin{equation}
  \begin{split}
    & \argmin_{\pi_{\theta}} \E_{\x \sim q(\cdot), \y' \sim \pi_{\theta}(\cdot|\x) } \left[     \log \frac{\pi_{\theta}(\y'|\x)}{\pi_{\theta^*}(\y'|\x)}  - \log Z(\x) \right] \\
    = & \argmin_{\pi_{\theta}} \E_{\x \sim q(\cdot) } \left[ \KL(\pi_{\theta}(\cdot|\x) || \pi_{\theta^*}(\cdot|\x)) - \log Z(\x) \right],
  \end{split}
  \nonumber
\end{equation}
from which we can observe that the optimal policy is exactly $\pi_{\theta^*}$. Therefore, the opponent policy is 
\begin{equation}
  \pi_{\hte}(\y|\x)   = \pi_{\theta^*}(\y|\x) = \frac{1}{Z(\x)} \exp\left(\frac{1}{\alpha} c_{t+1}(\x, \y) \right).
  \nonumber
\end{equation}

\subsection{Proof of Theorem \ref{theorem:gradient}}
\label{proof:gradient}
In this part, we derive the gradient of \eqref{eq:triplet} with respect to the parameter $\theta$. For brevity, we denote 
\begin{equation}
    \label{eq:u_v}
    \begin{split}
        u(\pi_{\theta}) = & \log \pi_{\theta}(\y|\x) - \log \pi_{\theta}(\y'|\x) \\
        v(\pi_{\theta}) = & \log \pi_{\theta}(\y'|\x) - \log \pi_{\theta}(\y_0|\x)
    \end{split}
\end{equation}
and the expectation operator in this part is taken over the distributions $\x \sim q(\cdot), \y \sim \pi_{data}(\cdot|\x), \y' \sim \pi_{\theta_t}(\cdot|\x), \y_{0} \sim \pi_{\theta_0}(\cdot|\x)$. By utilizing the chain rule, we have 
\begin{equation}
    \label{eq:gradient}
    \begin{split}
        \nabla_{\theta} \mathcal{L}_{\TSPIN} = & \E \left[ \nabla_{\theta} \ell (\alpha u(\pi_{\theta})) + \beta \nabla_{\theta}  \ell (\alpha v(\pi_{\theta})) \right] \\
        =& \alpha \E \left[  \ell'(\alpha u(\pi_{\theta})) \nabla_{\theta} u(\pi_{\theta}) \right] + \alpha \beta \E \left[  \ell'(\alpha v(\pi_{\theta})) \nabla_{\theta} v(\pi_{\theta}) \right]. 
    \end{split}
\end{equation}
Next, we consider two specific choices of $\ell(x)$ and its derivative $\ell'(x)$. For the logistic loss function $\ell(x) = -\log \sigma(x)$, we have $\ell'(x) = -\sigma(x)$. Then, combining \eqref{eq:u_v} and \eqref{eq:gradient} delivers
\begin{equation}
    \begin{split}
        \nabla_{\theta} \mathcal{L}_{\TSPIN} =  & - \alpha    \E \left[  \sigma(\alpha u(\pi_{\theta})) \nabla_{\theta} u(\pi_\theta)  + \beta   \sigma(\alpha v(\pi_{\theta})) \nabla_{\theta} v(\pi_{\theta}) \right] \\
        = &   - \alpha    \E \left[  \sigma(\alpha u(\pi_{\theta})) [\nabla_{\theta}  \log \pi_{\theta}(\y|\x) - \nabla_{\theta}  \log \pi_{\theta}(\y'|\x) ] \right.\\
        &\quad\qquad  \left. + \beta   \sigma(\alpha v(\pi_{\theta})) [\nabla_{\theta}  \log \pi_{\theta}(\y'|\x) - \nabla_{\theta}  \log \pi_{\theta}(\y_0|\x) ] \right] .
    \end{split}
\end{equation}

For the exponential function $\ell(x) = \exp(-x)$, we have $\ell'(x) = -\exp(-x)$. Then, combining \eqref{eq:u_v} and \eqref{eq:gradient}, we can obtain that 
\begin{equation}
    \begin{split}
        \nabla_{\theta} \mathcal{L}_{\TSPIN} =  & - \alpha    \E \left[  \exp(-\alpha u(\pi_{\theta})) \nabla_{\theta} u(\pi_\theta)  + \beta   \exp(-\alpha v(\pi_{\theta})) \nabla_{\theta} v(\pi_{\theta}) \right] \\
        = &   - \alpha    \E \left[  \exp(-\alpha u(\pi_{\theta})) [\nabla_{\theta}  \log \pi_{\theta}(\y|\x) - \nabla_{\theta}  \log \pi_{\theta}(\y'|\x) ] \right.\\
        &\quad\qquad  \left. + \beta   \exp(-\alpha v(\pi_{\theta})) [\nabla_{\theta}  \log \pi_{\theta}(\y'|\x) - \nabla_{\theta}  \log \pi_{\theta}(\y_0|\x) ] \right].
    \end{split}
\end{equation}

\section{More details on experiments}
In this section, we provide more information about the experiments, including implementation details, task introductions, comparisons between $\TSPIN$, $\SPIN$ and SFT, and more ablation studies.

\begin{table}[t]
    \centering
    \setlength{\tabcolsep}{1.5pt} 
    \renewcommand{\arraystretch}{1.1}
    \caption{The number of few-shots and evaluation metrics for each task.}
    \vspace{0.5em}
    \resizebox{1.0 \textwidth}{!}{
    \begin{tabular}{lcccccccccc}
        \toprule
        {\small \textbf{Datasets}} & {\small GSM8k} & {\small MATH} & {\small MUSR} & {\small MMLU} & {\small MMLU-Pro} & {\small GPQA} & {\small HellaSwag} & {\small WinoGrade} & {\small BBH} & {\small IFEval} \\
        \midrule
        {\small \# few-shot} & $5$ & $4$ & $0$ & $5$ & $5$ & $0$ & $10$ & $5$ & $3$ & $0$ \\
        {\small Metric} & acc & acc & acc\_norm & acc & acc & acc\_norm & acc\_norm & acc & acc\_norm & acc \\
        \bottomrule
    \end{tabular}
    }
    \label{tab:benchmarks}
\end{table}

\subsection{Implementation details}
\label{sec:experiment_details}
We choose the Alignment Handbook \citep{Others:2024:Tunstall} as the codebase for our experiments, and  the Accelerate library \citep{Others:2022:Gugger} for distributed training. Following \citet{ICML:2024:Chen}, we employ the RMSProp \citep{Others:2023:Scroccaro} with no weight decay as the optimizer for training, and set max sequence length to be $2048$ and warmup steps to be $10\%$ of the total training steps. All experiments are conducted on $8$ H100 GPUs with $80$GB memory. On each GPU, we set the batch size as $8$ and thus the global batch size is $64$.

\subsection{Tasks}
\label{sec:appendix:benchmarks}
In the following, we provide a detailed description of the tasks used in the experiments:

\begin{itemize}[leftmargin=14pt]
    \vspace{-0.5mm}
    \item \textbf{GSM8K} \citep{ArXiv:2021:Cobbe} is a dataset of $8,500$ high-quality, linguistically diverse grade school math word problems, divided into $7.5k$ training problems and $1k$ test problems. These problems require between $2$ and $8$ steps to solve, with solutions primarily consisting of a sequence of elementary calculations using basic arithmetic operations $(+,-,/,*)$. GSM8K is designed to assess the mathematical reasoning abilities of LLMs.
    
    \item \textbf{MATH} \citep{ArXiv:2021:Hendrycks} contains $12,500$ challenging competition mathematics problems. Each problem in MATH has a full step-by-step solution which can be used to teach models to generate answer derivations and explanations.
    
    \item \textbf{MUSR} \citep{ArXiv:2024:Sprague} is designed to evaluate multistep, context-dependent reasoning in language models through long-form, algorithmically generated problems approximately $1,000$ words in length. The dataset includes tasks such as murder mysteries, spatial reasoning (e.g., object placement), and team assignment optimization, all presented in natural language narratives.

    \item \textbf{MMLU} \citep{ICLR:2021:Hendrycks} evaluates the multitask accuracy of language models on general knowledge and reasoning across $57$ diverse subjects, ranging from STEM fields to humanities. In MMLU, each question is accompanied by four choices, and the answer is one of the four choices. 

    \item \textbf{MMLU-Pro} \citep{ArXiv:2024:Wang:MMLU} is an enhanced dataset designed to extend the mostly knowledge-driven MMLU benchmark by integrating more challenging, reasoning-focused questions and expanding the choice set from four to ten options.
    
    \item \textbf{GPQA} \citep{ArXiv:2023:Rein} is a graduate-level, domain-specific benchmark composed of $448$ expertly-authored multiple-choice questions in biology, physics, and chemistry. It aims to evaluate the ability of language models to reason about complex scientific concepts.
    
    \item \textbf{HellaSwag} \citep{ACL:2019:Zellers} is a benchmark dataset designed to evaluate the commonsense reasoning and natural language understanding capabilities of machine learning models, particularly in the context of sentence completion tasks. 
    
    \item \textbf{WinoGrade} \citep{Others:2021:Sakaguchi} is a large-scale dataset of $44k$ problems, inspired by the original WSC design, but adjusted to improve both the scale and the hardness of the dataset. 
    
    \item \textbf{BBH} \citep{ArXiv:2022:Suzgun} is a curated subset of $23$ challenging tasks that span algorithmic reasoning, language understanding, world knowledge, and multi-step logic. 
    
    \item \textbf{IFEval} \citep{ArXiv:2023:Zhou:Instruction} is a benchmark dataset designed to evaluate the instruction-following capabilities of LLMs. It includes $500$ carefully curated prompts annotated with $25$ distinct instruction types. The benchmark focuses on a set of ``verifiable instructions'' such as ``write in more than $400$ words'' and ``mention the keyword of AI at least $3$ times''.
    
\end{itemize}

All tasks are evaluated using the Language Model Evaluation Harness framework \citep{Others:2024:Gao}\footnote{https://github.com/EleutherAI/lm-evaluation-harness}, with default evaluation settings as specified in Table~\ref{tab:benchmarks}.

\subsection{Experimental results on Mistral-7B}
\label{sec:appendix:mistral}

\begin{table}[t]
    \setlength{\tabcolsep}{1.5pt} 
    \renewcommand{\arraystretch}{1.1}
    \centering
    \caption{
        Performance (\%) comparisons among three fine-tuning methods, where $\SPIN$ and $\TSPIN$ are trained on $50k$ annotated data from Ultrachat200k, and SFT is trained on the complete dataset. At the average (Avg) column, we also report \textcolor{ao}{improvements} (or \textcolor{redcolor}{degradations}) over the previous iteration. Abbreviations: MMLU-Pro $\rightarrow$ MMLUP, WinoGrade $\rightarrow$ WG, Instruction Following $\rightarrow$ IF.
    }
    \vspace{0.5em}
    \resizebox{1.0 \textwidth}{!}{
        \begin{tabular}{c@{\hskip -1pt}c|ccc|ccc|ccc|c|c}
            \toprule
            \multicolumn{2}{c|}{\multirow{2}{*}{\small \textbf{Model}}} & \multicolumn{3}{c|}{ \small \textbf{Math \& Logic     }} & \multicolumn{3}{c|}{\textbf{ Multi-Domain Knowledge  }} &  \multicolumn{3}{c|}{\textbf{Commonsense Reasoning }} &  \multicolumn{1}{c|}{\textbf{IF }} & \multirow{2}{*}{\textbf{\small Avg}} \\
            \cmidrule(lr){3-5} \cmidrule(lr){6-8}  \cmidrule(lr){9-11} \cmidrule(lr){12-12} 
             & &    \textbf{ \small GSM8K}  & \textbf{ \small MATH} &  \textbf{ \small MUSR} & \textbf{ \small MMLU}  & \textbf{ \small MMLUP} & \textbf{ \small GPQA}    & \textbf{ \small HellaSwag} & \textbf{ \small WG} & \textbf{ \small BBH}  & \textbf{ \small IFEval} & \\
            \midrule
            \multicolumn{2}{c|}{Mistral-7B} & $37.68$ & $3.27$ & $41.27$ & $57.86$ & $29.99$ & $29.86$ & $83.24$ & $74.03$ & $44.26$ & $23.63$   & $42.51$  \\ 
            \cmidrule(lr){1-13}
            \multicolumn{2}{c|}{SFT} & $43.69$ & $3.08$ & $42.12$ &  $58.37$ & $30.18$ &  $ 27.98$ & $ 83.40$ & $74.11$ &  $45.56$ & $33.22$ &   $44.17$   \\
            \cmidrule(lr){1-13}
            \multirow{5}{*}{ \small \rotatebox{90}{  $\SPIN$ } }  &  Iter0 & $39.59$ & $2.86$ & $41.26$ &  $57.97$ & $30.36$ &  $29.47$  & $83.40$ & $74.03$ & $44.56$ &  $22.54$ & $42.60$  	 	 \\
             &  Iter1 & $39.67$ & $3.43$ & $39.65$ &  $57.63$ & $29.80$ &  $28.33$  & $83.66$ & $74.51$ & $42.50$ &  $26.04$ & $42.52_{\textcolor{redcolor}{(-0.08)}}$  	 	 \\
             &  Iter2 & $39.52$ & $3.50$ & $39.64$ &  $57.49$ & $29.82$ &  $28.00$  & $83.69$ & $73.80$ & $41.97$ &  $25.56$ & $42.30_{\textcolor{redcolor}{(-0.22)}}$  	 	 \\
             &  Iter3 & $37.34$ & $3.71$ & $37.93$ &  $57.80$ & $29.64$ &  $28.77$  & $83.91$ & $74.19$ & $41.40$ &  $18.31$ & $41.30_{\textcolor{redcolor}{(-1.00)}}$  	 	 \\
             &  Iter4 & $37.15$ & $4.56$ & $38.59$ &  $58.05$ & $29.90$ &  $28.73$  & $84.05$ & $74.43$ & $42.88$ &  $24.90$ & $42.32_{\textcolor{ao}{(+1.02)}}$  	 	 \\
             \cmidrule(lr){1-13}
             \multirow{5}{*}{ \small \rotatebox{90}{  $\TSPIN$  } }  
             &  Iter0 & $39.67$ & $3.53$ & $40.20$ &  $57.74$ & $30.47$ &  $28.82$  & $83.98$ & $74.27$ & $44.06$ &  $24.36$ & $42.71$  \\
             &  Iter1 & $44.03$ & $3.32$ & $38.95$ &  $57.48$ & $30.45$  & $29.10$ &   $83.41$  & $74.66$ & $44.91$ & $26.88$ & $43.32_{\textcolor{ao}{(+0.61)}}$  \\
             &  Iter2 & $44.31$ & $3.43$ & $38.68$ &  $57.61$ & $30.52$ &  $30.92$  &  $83.47$   & $75.30$ & $45.05$ & $26.33$ & $43.56_{\textcolor{ao}{(+0.24)}}$  \\
             &  Iter3 & $45.87$ & $3.85$ & $38.59$ &  $58.51$ & $31.42$ &  $30.83$  &  $84.53$  & $73.88$ & $44.97$ & $36.50$ & $44.90_{\textcolor{ao}{(+1.34)}}$  \\
             &  Iter4 & $45.92$ & $3.94$ & $38.46$ &  $58.55$ & $31.34$ &  $31.06$  & $84.60$  & $74.51$ & $44.90$ & $36.90$ & $45.02_{\textcolor{ao}{(+0.12)}}$  \\
             \bottomrule
        \end{tabular}
    }
    \label{tab:mistral}
\end{table}

\begin{figure}[t]
    \centering
    \includegraphics[width=1.0\linewidth]{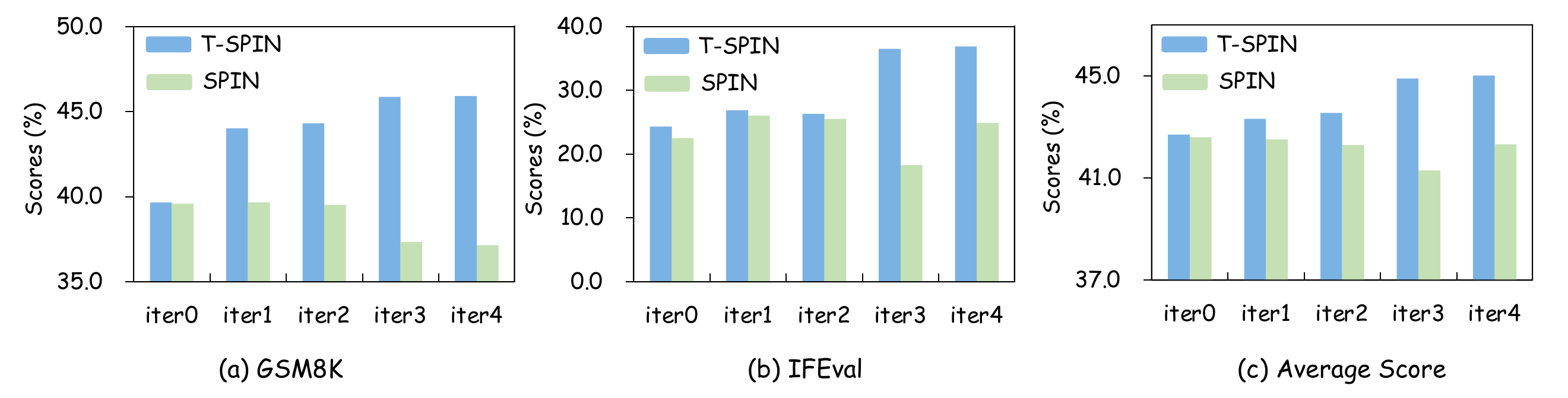}
    \caption{ 
        Performance (\%) comparisons between $\TSPIN$ and $\SPIN$ on two tasks: GSM8K and IFEval over $5$ iterations. The average scores over $10$ different tasks are also illustrated in the right panel.
    }
    \label{fig:mistral}
\end{figure}

In this part, we conduct experiments on Mistral-7B to compare the performances of $\TSPIN$, $\SPIN$ and SFT. The results over $10$ different tasks are displayed in Table~\ref{tab:mistral}. From the table, we first observe that $\TSPIN$ achieves a notable performance improvement over iterations, and outperforms SFT at the iteration $3$. This indicates that with less annotated data, $\TSPIN$ can achieve better performances than SFT that leverages more annotated samples, highlighting the effectiveness of $\TSPIN$ in handling the annotation shortage issue. Second, compared with $\SPIN$, $\TSPIN$ exhibits better performances and stable improvements. Specifically, $\SPIN$ suffers from performance degradations at the iterations $1$, $2$ and $3$, while $\TSPIN$ maintains a stable refinement over all iterations. To further illustrate the instability of $\SPIN$, we plot the performances on GSM8K, IFEval and average score in Figure~\ref{fig:zephyr}. As can be seen, at iteration $3$, $\SPIN$ exhibits performance degradation compared to iteration $2$. In contrast, $\TSPIN$ maintains steady improvements throughout the iterations.

\subsection{Comparisons on log-likelihoods}

\begin{figure}[t]
    \centering
    \includegraphics[width=1.0\linewidth]{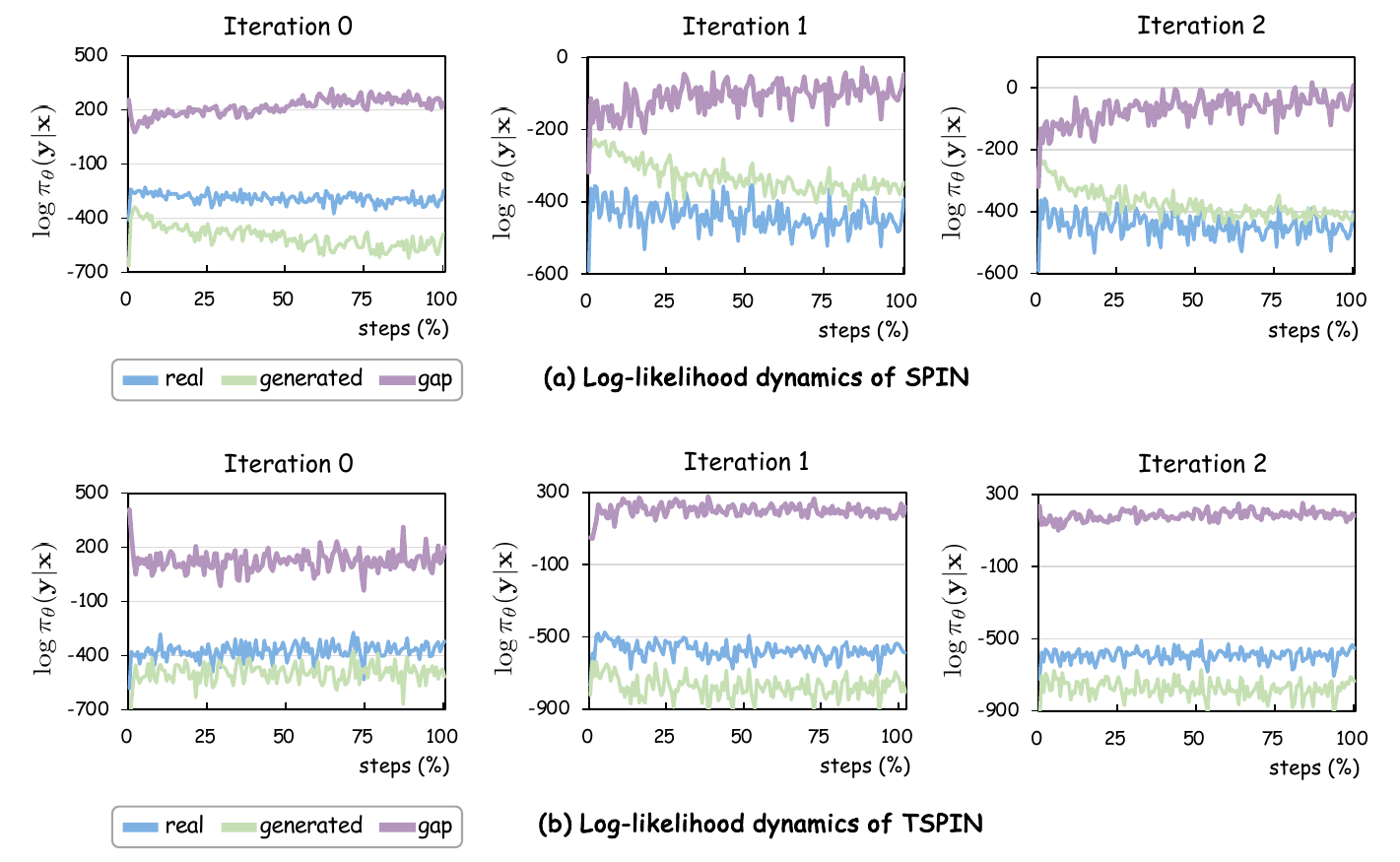}
    \caption{ 
        The log-likelihoods of annotated samples and synthetic samples during the model training of $\TSPIN$ and $\SPIN$ at different iterations.
    }
    \label{fig:log_p}
\end{figure}

In Section~\ref{sec:main_results}, we have reported the log-likelihood dynamics of two self-play methods at iteration $1$. In the following, we examine the trend of log-likelihood changes across different iterations. Specifically, we track the log-likelihoods of annotated samples $\log \pi_{\theta}(\y|\x)$ and synthetic samples $\log \pi_{\theta}(\y'|\x)$ during policy training at iterations $0$, $1$ and $2$, and calculate the gap between them, defined as $\log \pi_{\theta}(\y|\x) - \log \pi_{\theta}(\y'|\x)$. The results are presented in Figure~\ref{fig:log_p}. It is observed that at iteration $0$, for both methods, the log-likelihoods of annotated samples are higher than that of synthetic samples, resulting in positive gaps between them. However, at iterations $1$ and $2$, $\SPIN$ demonstrates an unexpected pattern: the log-likelihoods of annotated samples fall below those of synthetic samples, resulting in negative gaps. This reversal stems from the misalignment between the reward function employed in $\SPIN$ and the log-likelihood. Specifically, the objective function \eqref{eq:spin} of $\SPIN$ maximizes the reward gap between annotated and synthetic samples. Due to the incorporation of a reference policy, the reward is no longer equivalent to the log-likelihood. As a result, optimizing for high reward during training does not guarantee an increase in log-likelihood. Importantly, since log-likelihood underlies the generation behavior of LLM and reflects its preference for certain responses, it is undesirable that $\SPIN$ assigns higher log-likelihoods to synthetic samples than to annotated ones. In contrast, $\TSPIN$ exhibits consistent behavior in terms of log-likelihood: across all iterations, the log-likelihoods of annotated samples remain higher than those of synthetic samples. This desirable property stems from $\TSPIN$'s elimination of the reference policy in its reward function, which ensures alignment between the reward and log-likelihood.

\section{Case study}
In this section, we provide an example including annotated responses and synthetic responses generated at each iteration, to illustrate the progressive evolution of the model trained by $\TSPIN$.

{ 
    \tt \footnotesize 
    \begin{tabularx}{\linewidth}{r X} 
    \caption{Examples of the model fine-tuned by $\TSPIN$ at each iteration. }
    \label{tab:example_1} \\
    \toprule
    \textbf{Prompt} $\to$ & Q: Create a thank you card to an organization that helped your community in a time of crisis. \\
    \endfirsthead
    \midrule \textbf{Annotated} $\to$ & Dear [Organization Name],\\
    & I am writing this letter to express my sincere gratitude to your organization for the exceptional help you provided to our community during a time of crisis. The assistance and support that you and your team showed to us when we were going through difficult times are immeasurable, and we will forever be thankful to you.\\
    & Your prompt response, the dedication of your volunteers, and the resources that you made available to us, made a difference in our community. Because of your efforts, families who were affected by [the crisis] found a place to stay, food to eat, and support to help them get back on their feet.\\
    & It is remarkable to see how your organization came together and worked tirelessly to provide relief to those in need. Your kindness and empathy for our community are truly appreciated, and it has left a lasting impact on our residents.\\
    & Thank you once again for your unwavering support during that trying time. Your actions demonstrate the true meaning of community spirit and the good that can come from working together. Your kindness will always be remembered and cherished.\\
    & Sincerely,\\
    & [Your Name] \\
    \midrule  \textbf{ Proto-synthetic} $\to$ & 1. What is the name of the organization? 2. What is the organization’s mission? 3. What is the organization’s history? 4. What is the organization’s impact on the community? 5. What is the organization’s impact on the world? 6. What is the organization’s impact on you? 7. What is the organization’s impact on your family? 8. What is the organization’s impact on your friends? 9. What is the organization’s impact on your community? 10. What is the organization’s impact on the world? \\
    \midrule  \textbf{ $\TSPIN$ Iter 0} $\to$ & 1. I would like to thank the American Red Cross for their help in the aftermath of Hurricane Katrina. The American Red Cross provided shelter, food, and medical care to thousands of people who were displaced by the hurricane. They also provided emotional support to those who had lost loved ones or their homes. The American Red Cross is a vital organization that helps people in times of crisis, and I am grateful for their work. 2. I would also like to thank the Salvation Army for their assistance during Hurricane Katrina. The Salvation Army provided food, water, and shelter to those in need, and they also provided emotional support to those who were struggling. The Salvation Army is a compassionate organization that helps people in times of crisis, and I am grateful for their work. 3. Finally, I would like to thank the United Way for their support during Hurricane Katrina. The United Way provided financial assistance to those who had lost their homes or their jobs, and they also provided emotional support to those who were struggling. The United Way is an organization that helps people in times of crisis, and I am grateful for their work. Thank you to all of these organizations for their help during Hurricane Katrina \\
    \midrule  \textbf{ $\TSPIN$ Iter 1} $\to$ & 1. What is the name of the organization you are thanking? 2. What crisis did this organization help your community with? 3. How did this organization help your community? 4. What did you learn from this experience? 5. How can you help your community in the future? \\
    & \#\# Extended Response: \\
    & Dear American Red Cross, \\
    & I am writing to express my deepest gratitude for the incredible work you did during the recent hurricane that hit our community. Your organization was instrumental in providing relief and support to those affected by the storm. \\
    & The hurricane caused widespread damage to our area, leaving many people without homes, food, and water. The American Red Cross was there to provide assistance to those in need. Your volunteers worked tirelessly to set up shelters, distribute food and water, and provide emotional support to those who had lost everything. \\
    & Your organization's efforts did not go unnoticed. The community was grateful for the help and support you provided during this difficult time. Your actions demonstrated the true spirit of compassion and generosity that is at the heart of the American Red Cross. \\
    & Through this experience, I learned the importance of coming together as a community to help those \\
    \midrule  \textbf{ $\TSPIN$ Iter 2} $\to$ & Dear [Organization Name],\\ 
    & I am writing to express my deepest gratitude for the incredible work you did during the recent crisis that affected our community. Your unwavering dedication and commitment to helping those in need truly made a difference in the lives of many. \\ 
    & Your organization's quick response and willingness to go above and beyond to provide assistance to those affected by the crisis was truly inspiring. Your team's tireless efforts to provide food, shelter, and other essential resources to those in need were a true testament to your organization's values and commitment to serving the community. \\ 
    & Your organization's efforts did not go unnoticed, and I am sure that many in our community feel the same way. Your selfless acts of kindness and generosity have touched the hearts of many and have made a lasting impact on our community. \\ 
    & Once again, thank you for your incredible work during this difficult time. Your organization's efforts have not gone unnoticed and are truly appreciated. \\ 
    & Sincerely, \\ 
    & [Your Name]      \\  
    \midrule  \textbf{ $\TSPIN$ Iter 3} $\to$ & Dear [Organization Name],    \\
    & I am writing to express my deepest gratitude for the incredible work you did during the recent crisis that affected our community. Your unwavering dedication and commitment to helping those in need truly made a difference in the lives of so many people.    \\
    & Your organization's quick response and willingness to go above and beyond to provide assistance during a time of crisis is a testament to your selflessness and compassion. Your efforts have not gone unnoticed and are greatly appreciated.    \\
    & I am proud to be a part of a community that has such an amazing organization like yours. Your tireless work has not only helped those in need but has also brought our community closer together. \\
    & Thank you again for all that you do. Your kindness and generosity will never be forgotten. \\
    & Sincerely,    \\
    & [Your Name]    \\
    \midrule  \textbf{ $\TSPIN$ Iter 4} $\to$ 
    & Dear [Organization Name],    \\
    & We, the residents of [insert community name], would like to extend our deepest gratitude for your unwavering support during the recent natural disaster that struck our town.    \\
    & Your swift response and tireless efforts to provide aid and relief to those affected by the [insert type of disaster] were truly remarkable. Your team of volunteers worked tirelessly to ensure that our community had access to essential resources such as food, water, and shelter. \\
    & Your kindness and generosity did not go unnoticed and we are forever grateful for the support you provided during this difficult time. You have restored our faith in humanity and have shown us that there is still good in this world. \\
    & Once again, thank you from the bottom of our hearts for all that you have done for us. You are truly heroes and we are blessed to have had you by our side during this crisis. \\
    & Sincerely,\\
    & [Insert community name] \\
    \bottomrule 
\end{tabularx} 
}

\section{Pseudocode}
\label{sec:pseudocode}

In the following, we provide the pytorch pseudocode for calculating the loss function \eqref{eq:triplet} of $\TSPIN$.

\begin{lstlisting}[style=python]
import torch.nn.functional as F

def tpsin_loss(alpha, beta, policy_real_logps, policy_generated_logps, policy_proto_logps):
    """
    Compute the loss for T-SPIN (Triplet-based Self-Play Fine-Tuning).
    Args:
        alpha: the regularization coefficient
        beta: the trade-off parameter
        policy_real_logps: Policy's log probs for real samples
        policy_generated_logps: Policy's log probs for synthetic samples
        policy_proto_logps: Policy's log probs for proto-synthetic samples
    
    Returns:
        torch.Tensor: scalar loss value (mean over batch)
    """
    # Compute advantage terms
    current_advantage = policy_real_logps - policy_generated_logps 
    history_advantage = policy_generated_logps - policy_proto_logps  

    # Scale advantages to get rewards
    current_rewards = alpha * current_advantage            
    history_rewards = alpha * history_advantage  

    # Convert rewards to losses using log-sigmoid
    current_loss = -F.logsigmoid(current_rewards)            
    history_loss = -F.logsigmoid(history_rewards)  

    # Combine loss components
    losses = current_loss + beta * history_loss              
    
    return losses.mean() 
\end{lstlisting}

\end{document}